\documentclass[11pt]{article}
\usepackage{amsmath}
\usepackage{amssymb}
\usepackage[final]{acl}

\usepackage{times}
\usepackage{latexsym}
\usepackage{tcolorbox}
\tcbuselibrary{skins}
\usepackage{booktabs}
\usepackage{url}

\usepackage[T1]{fontenc}
\DeclareUnicodeCharacter{00A0}{ }

\usepackage[utf8]{inputenc}

\usepackage{microtype}

\usepackage{inconsolata}

\usepackage{graphicx}
\usepackage{tabularx}

\title{Divide-Prompt-Refine: a Training-Free, Structure-Aware Framework for Biomedical Abstract Generation}

\author{
 \textbf{Sylvey Lin\textsuperscript{1}},
 \textbf{Joe Menke\textsuperscript{1}},
 \textbf{Shufan Ming \textsuperscript{1}},
 \textbf{Dongin Nam\textsuperscript{1}},
\\
 \textbf{Neil Smalheiser\textsuperscript{1,2}},
 \textbf{Halil Kilicoglu\textsuperscript{1}},
\\
\\
 \textsuperscript{1} School of Information Sciences, University of Illinois Urbana-Champaign, Champaign, IL\\
 \textsuperscript{2} Department of Psychiatry, University of Illinois College of Medicine, Chicago, IL
\\
 \small{
   \textbf{Correspondence:} \href{mailto:yuhsinl2@illinois.edu}{yuhsinl2@illinois.edu}
 }
}

\begin{document}
\maketitle
\begin{abstract}
Biomedical abstracts play a critical role in downstream NLP applications, such as information retrieval, biocuration, and biomedical knowledge discovery. However, a non-trivial number of biomedical articles do not have abstracts, diminishing the utility of these articles for downstream tasks. We propose DPR-BAG (Divide, Prompt, and Refine for Biomedical Abstract Generation), a training-free, zero-shot framework that generates coherent and factually grounded abstracts for biomedical articles with full text but no abstract. DPR-BAG decomposes full-text documents into structured rhetorical facets following the Background-Objective-Methods-Results-Conclusions (BOMRC) schema, performs parallel LLM-based summarization for each facet, and applies a final refinement stage to restore global discourse coherence. On PMC-MAD, a distribution-aligned dataset of 46,309 biomedical articles, DPR-BAG improves abstractive novelty over strong extractive and fine-tuned baselines, while maintaining factual consistency. Our ablation study reveals a counterintuitive finding: increasing prompt complexity or explicitly injecting entity-level guidance can degrade factual alignment, highlighting the importance of controlled prompting strategies. These findings underscore the potential of training-free, structure-aware frameworks for scalable biomedical abstract generation in low-resource settings. Our data and code are available at 
\url{https://huggingface.co/datasets/pmc-mad/PMC-MAD} and \url{https://github.com/ScienceNLP-Lab/MultiTagger-v2/tree/main/DPR-BAG}.

\end{abstract}

\section{Introduction}
Many biomedical NLP tasks rely heavily on abstracts, due to their accessibility and information density. Abstracts provide an author-written summary of core scientific findings, making them a useful proxy for full-text articles in downstream applications. For example, \citet{ir1} showed that pre-training tasks designed around the title-abstract structure improve biomedical information retrieval; \citet{biocuration1} used abstracts as an initial data source for biocuration. Beyond content, the structured organization of abstracts also benefits downstream tasks: \citet{ir2} leverage abstract-level structure to refine retrieval; PubMedQA \cite{pubmedQA} is based on structured abstracts to support high-fidelity biomedical knowledge discovery. 
Moreover, abstracts alone can serve as a stronger training signal than full texts in some settings \cite{pubmedBERT}.

However, the absence of abstracts in a significant portion of biomedical articles creates a bottleneck for these tasks. 
As of April 2026, 11,603,796 out of 40,414,072 (\textasciitilde29\%) PubMed articles were missing abstracts, with this volume continuing to rise despite a declining overall proportion, driven by the growth of publication types such as case reports, editorials, and letters. These publication types carry substantial scientific value. For instance,  \citet{gurulingappa2012development} and  \citet{case_report} utilized case reports for adverse drug event detection; \citet{magnet2006letters} and \citet{letters} assessed letters to characterize post-publication scientific discourse, including patterns of critique, rhetorical features of disagreement, and trends in authorship; and \citet{Waaijer2011} and \citet{ioannidis2025house} analyzed editorials to study how journals shape scientific discourse, including the distribution of topics, the framing of policy issues, and the presence of systematic biases.

The absence of abstracts in these articles motivates the Biomedical Abstract Generation (BAG) task, which aims to automatically generate abstracts from full-text biomedical articles. Although related to standard document summarization, BAG differs in important ways. It must adhere to scientific reporting conventions, including structured presentation of methods, results, and conclusions, while preserving fine-grained biomedical entities, quantitative findings, and explicit argumentative relationships that are often critical for scientific interpretation. Early BAG work by \citet{chachra2016} utilized extractive sentence selection, which can lead to fragmented coherence and poor lexical flow. Moreover, because full-length biomedical articles often exceed the context limits of standard models, BAG is inherently a long-context task, making it vulnerable to extractive bias and factual fidelity issues. 
For example, \citet{scholarSum} demonstrates that even state-of-the-art models like GPT-4 suffer from hallucinations and information omission when extracting from non-decomposed scientific full texts, emphasizing the inherent fidelity risks in long-document processing required for BAG task. Beyond fidelity risks, recent analysis also reveals that when forced to process complex long full texts, even specialized models like LongT5 exhibit a strong extractive bias, relying on simple heuristics to copy verbatim snippets rather than synthesizing information \cite{abstractiveSum}. As a result, these models can suffer from the same core issue as traditional extractive summarizers: they produce fragmented text that lacks the natural flow and cohesion of human-written summaries \cite{a_vs_e}.

To address these limitations, we propose the Divide, Prompt and Refine Biomedical Article Generation (DPR-BAG) framework. Drawing on prior work showing that divide-and-conquer decomposition reduces intermediate errors in LLMs \cite{splittingmakesstronger}, DPR-BAG decomposes full-text articles along their rhetorical structure, performs parallel summarization on each resulting facet, and applies a modular refinement stage to reconcile fragmented outputs and restore discourse coherence. We target six rhetorical facets: Background, Objectives, Methods, Results, Conclusions (BOMRC), and Others. BOMRC is adopted as it represents the standard discourse structure validated in the PubMed 200k RCT dataset \cite{pubmed200krct}, while the ``Others" facet retains any unclassified content.
Using this design, we focus on three research questions: 
\begin{enumerate}
    \item Can we develop a training-free approach for the BAG task? 
    \item Does structure-aware decomposition of full-text articles improve the quality of generated abstracts compared to naive prompting?
    \item To what extent does increasing prompting complexity (from detailed instructions to entity guidance) improve generation quality? 
\end{enumerate}

Our main contributions are as follows:
\begin{enumerate}
    \item We propose DPR-BAG, a training-free, structure-aware method for BAG. 
    \item We release a dataset of more than 46K biomedical full-text publications for BAG task.
    \item We compare DPR-BAG to strong extractive and abstractive baselines.
    \item We systematically evaluate the effect of various prompting and splitting strategies as well as entity guidance within DPR-BAG. 
\end{enumerate}

\begin{figure*}[t]
    \centering
    \includegraphics[width=\textwidth]{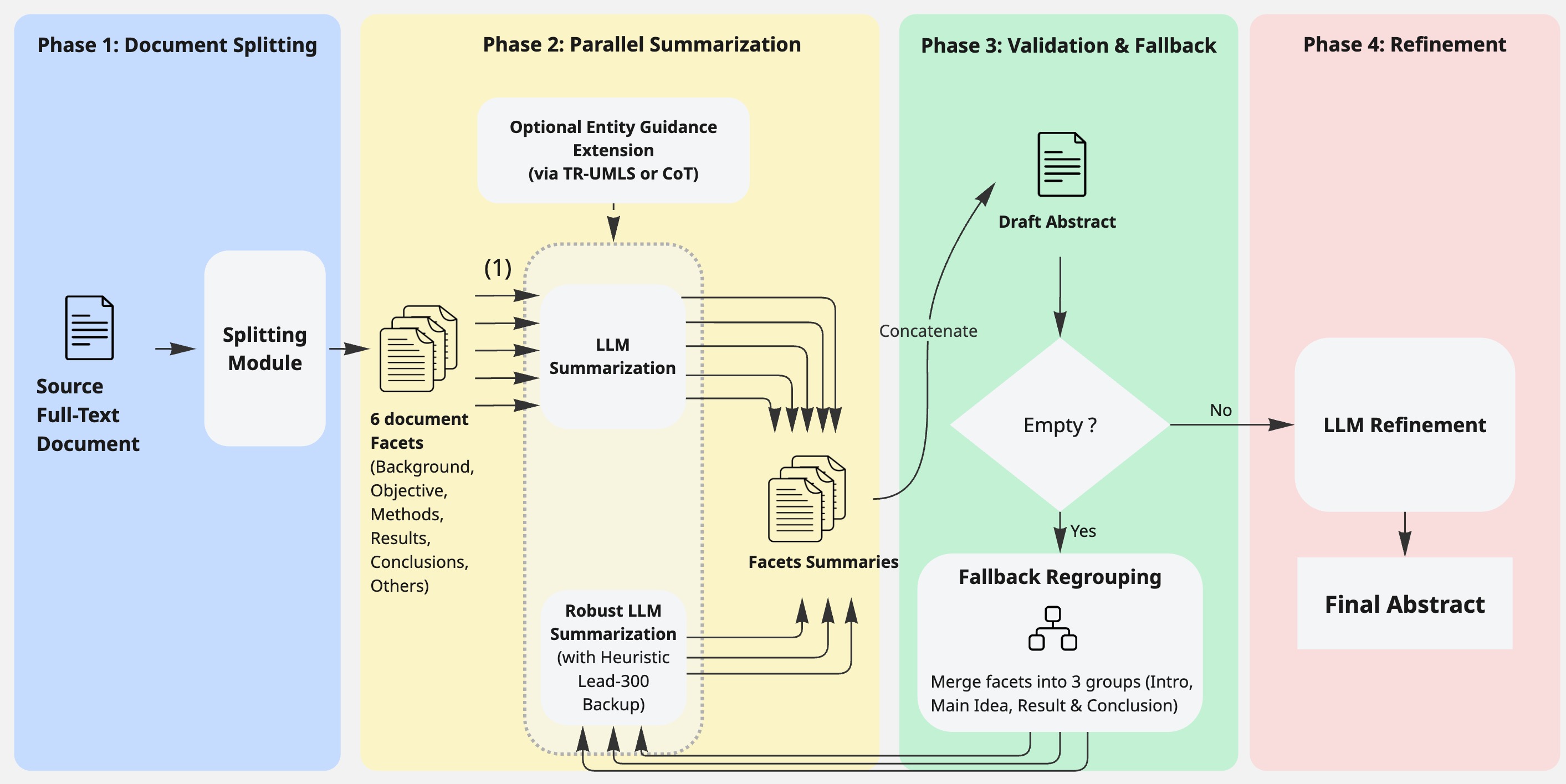} 
    \caption{Overview of the DPR-BAG framework for biomedical abstract generation.}
    \label{fig:framework}
\end{figure*}
\section{Dataset}
We constructed a BAG dataset based on PubMed publications from 1987 to 2023. To ensure a representative sample, we first calculated the publication type (PT) distribution of articles lacking abstracts using PT queries adapted from prior work \cite{multitagger}. We then performed stratified sampling based on this distribution to retrieve 130,000 candidate XML files from the PubMed Central (PMC) Open Access subset, ensuring that the sampled articles reflect the publication type distribution of abstract-less PubMed records. For data processing, we adapted the extraction pipeline from the Long-summarization framework \cite{pubmed_dataset} to parse structured sections and abstracts from the raw XML files. After filtering out records that were unparseable or lacked extractable abstracts, the final dataset, hereafter referred to as PMC-MAD (Missing-Abstract Distribution-aligned PMC), consists of 46,309 articles.


\section{Methods}
DPR-BAG follows a modular pipeline designed to generate structure-aware abstracts from biomedical full-text articles (Figure \ref{fig:framework}). The process begins by decomposing the document into five distinct facets based on BOMRC, plus an "Others" facet for unclassified content.
For each facet, we perform parallel LLM-based summarization, which can optionally be augmented with entity guidance extension. The resulting sectional summaries are then concatenated and passed to a final LLM-based refinement stage to restore discourse coherence. DPR-BAG requires no task-specific training or fine-tuning; all components operate in a zero-shot manner using pre-trained, off-the-shelf models.

\subsection{Task Formulation}
Given a full-text biomedical article $D = (p_1, p_2, \ldots, p_n)$ , where each $p_i$ denotes a paragraph, the goal is to generate an abstract $A$ that covers a predefined set of rhetorical facets $\mathcal{F} = \{$Background, Objectives, Methods, Results, Conclusions, Others$\}$ (BOMRC+).

We reformulate this as a facet-conditioned summarization problem, where the model generates content for each rhetorical facet separately. This decomposition allows the model to address each rhetorical component independently and capture discourse structure.
Specifically, the document is partitioned into $K = 6$ facet-specific sub-documents $\{D_{f_k}\}_{k=1}^{K}$, where each $D_{f_k}$ aggregates paragraphs rhetorically aligned with facet $f_k$. Each sub-document is independently summarized to produce a facet summary $\hat{a}_{f_k}$, and the concatenation $\hat{A} = \bigoplus_{k=1}^{K} \hat{a}_{f_k}$ is subsequently refined into the final abstract $R(\hat{A}) = A$. Facets absent from the source document yield empty strings. 

\subsection{Document Splitting} 
To segment full-text documents into rhetorically coherent texts, we use LLM-SSC \cite{LLM-SSC}, an LLM-based sequential sentence classification framework that assigns rhetorical labels (BOMRC) to sentences using in-context learning (performance details in Appendix~\ref{appendix:LLM-SSC}). While the model was trained on structured abstracts, we assume that the underlying rhetorical intent (e.g., methodological description vs. result reporting) remains consistent within full-text paragraphs. Specifically, we leverage the role of the first sentence of the paragraph as a topic sentence that typically encapsulates the paragraph's functional purpose. We assign the label of the first sentence of each paragraph as the global label for that paragraph, subsequently concatenating all paragraphs with matching labels to form the input document facets. We refer to this approach as the \textbf{First Sentence Labeling (FS)} strategy, and empirically validate it against naive splitting (NS) and section-header (SH) ablation variants below.

\paragraph{Naive Splitting Approach (NS):} This approach distributes paragraphs into six segments, aiming for an approximately even distribution while maintaining paragraph integrity. Including this baseline allows us to assess whether a semantic-aware division (e.g., LLM-SSC) offers advantages over a purely structural, length-based partition.
\paragraph{Section-header Normalization (SH):} This strategy serves as a coarse-grained semantic baseline. Utilizing the Transformer model developed in \citet{secnormposter}, this approach categorizes existing section headers into standard BOMRC categories and concatenates paragraphs within the same facet. This comparison helps determine if the fine-grained, sentence-level classification used in LLM-SSC provides additional utility beyond simple section-level organization.

\subsection{Parallel Summarization}
After the documents are divided into six document facets (BOMRC and Others), each is input into the LLM to generate a corresponding facet summary; facets that are not present in the source document are represented as empty summaries. 
These summaries are subsequently concatenated to form the draft abstract. The following subsections detail the prompting strategies and optional entity guidance extensions used during summarization.

\subsubsection{Prompting Strategies}
To investigate the effect of prompt complexity on generation quality,
we adopt a \textbf{Basic Concise (BC)} prompting strategy as the baseline, and ablate prompting complexity by evaluating two more elaborate variants, \textbf{Detailed Instruction (DI)} and \textbf{Structural Instruction (SI)}. Full prompt templates are described in Appendix~\ref{appendix:prompts}.

\paragraph{Basic Concise Prompting (BC):} BC is a minimal prompting strategy with coarse-grained focal points for each rhetorical facet (e.g., directing the model to ``prioritize key findings and data'' for the Results section) without further elaboration or explicit formatting structure. 

\paragraph{Detailed Instruction Prompt (DI):} DI is a more detailed prompting strategy modeled after the abstract submission guidelines of JMIR Publications\footnote{\url{https://support.jmir.org/hc/en-us/articles/37982552280987-Submitting-Your-Manuscript-to-JMIR-Publications-A-Guide-for-Authors}\label{footnote:jmir}} whose five-part BOMRC structured guideline aligns with the target rhetorical categories that DPR-BAG uses. 
By shifting the LLM persona to a ``biomedical synthesis assistant,'' this prompt aims to enforce the extraction of granular details and mandates the inclusion of specific research designs, sample sizes, response rates, and statistical metrics (such as p-values and confidence intervals) to ensure adherence to multifaceted reporting standards.

\paragraph{Structural Instruction Prompt (SI):} SI extends the basic prompt by introducing an explicit structural schema using Markdown formatting strategy, inspired by \citet{formattingimpact}. Compared to DI, which focuses on detailed content guidance, SI organizes the prompt into a more structured format, which aims to improve instruction adherence.


\subsubsection{Entity Guidance} We additionally introduce an optional knowledge-grounding extension to further enhance semantic fidelity during parallel summarization. This component extracts key biomedical entities to guide the LLM in summarizing each facet. We consider two instantiations of this extension, TR-UMLS and CoT, detailed below. 

\paragraph{TextRank and UMLS normalization (TR-UMLS):} We extract key phrases from each facet using TextRank, a graph-based unsupervised method that requires no additional training and is well-suited to our zero-shot setting, and link them to UMLS \cite{umls} concepts using scispaCy's UMLS entity linker \cite{scispacy}, retaining only phrases with valid UMLS mappings. When multiple phrases map to the same UMLS concept, they are grouped together and represented by a single term to avoid redundant anchoring. The top-$n$ entities, ranked by TextRank centrality, are incorporated into the summarization prompt as anchor terms to guide the model toward the most structurally significant medical information in the text. We ablate $n\in\{5,10\}$ in Section~\ref{sec:ablation}.

\paragraph{Chain-of-Thought (CoT):} As an alternative, we employ a two-stage prompting strategy within the LLM summarization module when the extension is activated. In the first stage, we prompt the model to list the important entities of the facet, and in the second stage, the model is prompted to synthesize the information based on the facet and the entities it listed in the first stage.\\

We ablate these two entity guidance strategies independently. TR-UMLS pairs with DI, while CoT pairs with SI, whose structured format makes it well-suited for CoT’s two-stage reasoning.

\subsection{Validation and Fallback} \label{sec:fallback}
To ensure the summarization pipeline robustness, we implement a validation and fallback mechanism. If all six facet summaries are empty due to insufficient context or LLM output format violations, the facets are regrouped into three broader categories (Intro, Main Idea, and Results \& Conclusions) and sent back to the parallel summarization module. If regrouping fails, a first 300 characters (Lead-300) heuristic backup is used to guarantee non-empty output (details in Appendix \ref{appendix:fallback}). 



\subsection{Refinement}
Upon the formation of the draft abstract, we prompt the LLM to perform a global refinement to ensure structural coherence and stylistic consistency. This final processing step synthesizes the concatenated facets into a unified Final Abstract (the refinement prompt is detailed in Appendix \ref{appendix:refined_prompt}).

\begin{table*}[t]
\centering
\small
\setlength{\tabcolsep}{5pt}
\begin{tabular}{lccc|ccc}
\hline
 & \multicolumn{3}{c}{\textbf{Abstractiveness}} & \multicolumn{3}{c}{\textbf{Factuality}}\\
 \cline{2-4} \cline{5-7}
\textbf{Model} & \textbf{Bi-g} & \textbf{Tri-g} & \textbf{Dens.} & \textbf{AS} & \textbf{MC} & \textbf{SummaC}\\
\hline
Original abstract & 0.495 & 0.663 & 6.650 & -- & -- & --\\
\hline
LED-Arxiv (Base) & 0.140 (-0.355) & 0.222 (-0.441) & 22.698 (+16.048) & 0.720 & 0.856 & \textbf{0.735}\\
LED-Pubmed (Base) & 0.167 (-0.328) & 0.261 (-0.402) & 19.233 (+12.583) & 0.667 & 0.845 & 0.693\\
LED-Pubmed (FT) & 0.309 (-0.187) & 0.453 (-0.210) & 11.369 (+4.719) & 0.511 & 0.653 & 0.503\\
LongT5 (Base) & 0.174 (-0.321) & 0.280 (-0.383) & 15.492 (+8.842) & 0.668 & 0.875 & 0.711\\
LongT5 (FT) & 0.252 (-0.243) & 0.379 (-0.284) & 12.989 (+6.339) & 0.540 & 0.725 & 0.565\\
\hline
\textbf{DPR-BAG} & \textbf{0.397 (-0.098)} & \textbf{0.605 (-0.058)} & \textbf{4.690 (-1.960)} & \textbf{0.762} & \textbf{0.890} & 0.570\\
\hline
\end{tabular}
\caption{Performance comparison on primary evaluation dimensions. For Abstractiveness metrics, parenthetical values indicate the difference from the original abstract; best = smallest absolute deviation. Factuality scores are computed against the source full-text; higher is better. Best results bolded. (Bi-g: Bigram novelty; Tri-g: Trigram novelty; Dens.: Density; AS: AlignScore; MC: MiniCheck)}
\label{tab:eval_primary}
\end{table*}

\begin{table*}[t]
\centering
\small
\setlength{\tabcolsep}{5pt}
\begin{tabular}{lccc|cccc}
\hline
 & \multicolumn{3}{c}{\textbf{Semantic Alignment}} & \multicolumn{4}{c}{\textbf{Supporting}}\\
 \cline{2-4} \cline{5-8}
\textbf{Model} & \textbf{BS} & \textbf{SENT} & \textbf{FOCUS$\downarrow$} & \textbf{R-L} & \textbf{U-R} & \textbf{Cov.} & \textbf{Comp.}\\
\hline
Original abstract & -- & -- & -- & -- & -- & 0.743 & 23.827\\
\hline
LED-Arxiv (Base) & 0.635 & 0.907 & 0.702 & 0.234 & 0.340 & 0.950 (+0.217) & 26.916 (+3.089)\\
LED-Pubmed (Base) & 0.652 & \textbf{0.908} & \textbf{0.670} & 0.262 & 0.354 & 0.944 (+0.201) & 27.387 (+3.560)\\
LED-Pubmed (FT) & 0.634 & 0.868 & 0.791 & 0.264 & \textbf{0.364} & \textbf{0.868 (+0.125)} & \textbf{20.901 (-2.926)}\\
LongT5 (Base) & \textbf{0.664} & 0.876 & 0.861 & 0.274 & 0.312 & 0.948 (+0.205) & 46.449 (+22.622)\\
LongT5 (FT) & 0.650 & 0.895 & 0.713 & \textbf{0.277} & 0.361 & 0.911 (+0.168) & 27.002 (+3.175)\\
\hline
\textbf{DPR-BAG} & 0.617 & 0.868 & 0.828 & 0.183 & 0.266 & 0.894 (+0.151) & 50.037 (+26.210)\\
\hline
\end{tabular}
\caption{Performance comparison on semantic alignment and supporting metrics. For Coverage and Compression, parenthetical values indicate the difference from the original abstract; best = smallest absolute deviation. BS, SENT, R-L, and U-R: higher is better. FOCUS: lower is better. Best results bolded. (BS: BERTScore F1; SENT: DS\_SENT\_NN; FOCUS: DS\_FOCUS\_NN; R-L: ROUGE-L; U-R: UMLS Recall; Cov.: Coverage; Comp.: Compression)}
\label{tab:eval_secondary}
\end{table*}
\section{Experimental Setup}
DPR-BAG is implemented with \texttt{Llama-3.2:3B} deployed via Ollama in instruction-tuned mode. Hardware details, token-limit constraints, and fine-tuning hyperparameters are provided in Appendix~\ref{appendix:implementation_details}.

\subsection{Baseline Models}
To establish robust baselines for our framework, we compared our approach against several standard long-document summarization models. We utilized two off-the-shelf variants of the Longformer Encoder-Decoder (LED) architecture \cite{beltagy2020longformer}---pretrained on arXiv\footnote{\url{https://huggingface.co/allenai/led-large-16384-arxiv}} and PubMed\footnote{\url{https://huggingface.co/patrickvonplaten/led-large-16384-pubmed}}, respectively---to evaluate their zero-shot transferability to our task. Additionally, we included an off-the-shelf LongT5 model \cite{longt5} pretrained on PubMed\footnote{\url{https://huggingface.co/Stancld/longt5-tglobal-large-16384-pubmed-3k_steps}} to broaden our baseline comparisons across different architectures. Finally, to ensure maximal adaptation to our corpus, we evaluated a supervised fine-tuned version of the PubMed-pretrained LED and LongT5 using our 80\% training split, using the 10\% validation split for early stopping.
All evaluations were performed on the remaining 10\% (test split). 




\subsection{Evaluation Metrics} To assess the generated abstracts, we employ a multi-dimensional suite of metrics, with particular emphasis on abstractiveness and factuality. Detailed formulations and implementation details of each metric are provided in Appendix \ref{appendix:metrics}. Paired bootstrap significance tests for the main comparisons are provided in Appendix \ref{appendix:significance_testing}.

\paragraph{Abstractiveness:} \textbf{Bigram and trigram novelty} measure the proportion of tokens absent from the source text, serving as a proxy for abstractive synthesis. We additionally adopt \textbf{Density} from Newsroom~\citep{newsroom}, which quantifies the length of verbatim copying from the source, offering a complementary view of the model's extractive behavior. For both, smaller absolute deviations from the human-written reference indicate closer stylistic alignment.

\paragraph{Factuality:} Given that our target abstracts are expected to be highly abstractive, factuality metrics must remain reliable under heavy paraphrasing. We therefore adopt \textbf{AlignScore}~\citep{alignscore} as our primary factuality measure, as it evaluates alignment against the source full-text across a broad range of dimensions (notably paraphrasing), making it more robust than purely entailment-based alternatives. We additionally report \textbf{SummaC}~\citep{summaC} and \textbf{MiniCheck}~\citep{minicheck}, both NLI-based metrics, as cross-checks.

\paragraph{Semantic Alignment:} We adopt \textbf{BERTScore} \citep{bertscore} to evaluate semantic similarity between generated abstracts and the reference abstracts via contextual embeddings. To further assess discourse coherence, we adopt \textbf{DiscoScore}~\citep{discoscore}, reporting DS\_SENT\_NN (sentence-level structural alignment) and DS\_FOCUS\_NN (shared noun semantic alignment).

\paragraph{Supporting Metrics:} We additionally adopt \textbf{ROUGE-L}~\citep{rouge} to measure n-gram overlap with the reference. \textbf{UMLS Recall} quantifies the proportion of UMLS~\citep{umls} concepts from the original abstract retained in the generated output. \textbf{Coverage} and \textbf{Compression} from Newsroom~\citep{newsroom} serve as complementary metrics to measure how source content is preserved and condensed. For these two metrics, as well, smaller absolute deviations from the human-written reference are preferred.


\begin{table*}[t]
\centering
\small
\begin{tabular}{lcc|cc|cc|c}
\hline
 & \multicolumn{2}{c}{\textbf{Abstractiveness}} & \multicolumn{2}{c}{\textbf{Factuality}} & \multicolumn{2}{c}{\textbf{Semantic Alignment}} & \\
 \cline{2-3} \cline{4-5} \cline{6-7} \cline{8-8}
\textbf{Splitting} & \textbf{Tri-g} & \textbf{Dens.} & \textbf{AS} & \textbf{MC} & \textbf{BS} & \textbf{FOCUS$\downarrow$} & \textbf{U-R}\\
\hline
FS & 0.605 (-0.058) & \textbf{4.690 (-1.960)} & \textbf{0.762} & 0.890 & 0.617 & \textbf{0.828} & \textbf{0.266}\\
NS & \textbf{0.613 (-0.050)} & 4.508 (-2.142) & 0.756 & \textbf{0.892} & 0.615 & 0.856 & 0.262\\
SH & 0.746 (+0.083) & 2.885 (-3.765) & 0.636 & 0.791 & \textbf{0.636} & 0.920 & 0.229\\
\hline
\end{tabular}
\caption{Ablation on document splitting strategies. Splitting strategies are Naive Splitting (NS), Section Header Normalization (SH), and First Sentence Labeling (FS). (Abbreviations as in Table~\ref{tab:eval_primary} and Table~\ref{tab:eval_secondary}) }
\label{tab:eval_splitting}
\end{table*}

\begin{table*}[t]
\centering
\small
\begin{tabular}{lcc|cc|cc|c}
\hline
 & \multicolumn{2}{c}{\textbf{Abstractiveness}} & \multicolumn{2}{c}{\textbf{Factuality}} & \multicolumn{2}{c}{\textbf{Semantic Alignment}} & \\
 \cline{2-3} \cline{4-5} \cline{6-7} \cline{8-8}
\textbf{Prompt} & \textbf{Tri-g} & \textbf{Dens.} & \textbf{AS} & \textbf{MC} & \textbf{BS} & \textbf{FOCUS$\downarrow$} & \textbf{U-R}\\
\hline
BC & \textbf{0.605 (-0.058)} & \textbf{4.690 (-1.960)} & \textbf{0.762} & \textbf{0.890} & 0.617 & \textbf{0.828} & \textbf{0.266}\\
DI & 0.725 (+0.063) & 3.131 (-3.519) & 0.642 & 0.806 & \textbf{0.638} & 0.885 & 0.247\\
SI & 0.736 (+0.073) & 3.049 (-3.601) & 0.630 & 0.805 & 0.636 & 0.836 & 0.251\\
\hline
\end{tabular}
\caption{Ablation on summarization prompt strategies. Prompt variants are Basic Concise (BC), Detailed Instruction (DI), and Structural Instruction (SI). (Abbreviations as in Table~\ref{tab:eval_primary} and Table~\ref{tab:eval_secondary})}
\label{tab:eval_prompts}
\end{table*}


\section{Results}
\label{sec:results}
\subsection{RQ1: Effectiveness of the Training-Free Approach}
We first evaluate whether the proposed training-free approach can achieve competitive performance on the BAG task. As shown in Table~\ref{tab:eval_primary} and Table~\ref{tab:eval_secondary}, DPR-BAG (BC prompt, no entity guidance extension) achieves competitive performance against fine-tuned baselines. It generates abstracts that more closely resemble human-written abstracts (more abstractive), while maintaining factual consistency with the full text (Table~\ref{tab:eval_primary}).

DPR-BAG outperforms the fine-tuned baselines in both abstractiveness and factuality (all paired bootstrap $p<0.001$ on Bigram and Trigram novelty, Density, AlignScore, and MiniCheck; Appendix \ref{appendix:sig_baselines}), while the fine-tuned baselines themselves perform better than other baselines in abstractiveness but their generations are less factual.

DPR-BAG yields mostly lower scores for semantic alignment and other supporting metrics compared to the baselines, reflecting a known bias in these metrics toward extractive outputs (Table~\ref{tab:eval_secondary}). Baseline models exhibit high density and low novelty relative to human-written abstracts (Table~\ref{tab:eval_primary}), indicating extractive behavior that likely inflates their scores. At the same time, DPR-BAG yields the highest compression rate (50.037), potentially leading to over-simplification of key information. We provide qualitative examples of generated abstracts in Appendix~\ref{appendix:case_study}.

\vspace{-0.25cm}
\subsection{RQ2: Impact of Structure-Aware Decomposition}
\label{sec:ablation}
We next evaluate whether structure-aware decomposition improves generation quality compared to naive prompting. 
As shown in Table \ref{tab:eval_splitting}, FS achieves the best overall performance. Compared to NS, FS achieves significantly higher AlignScore (paired diff = +0.006, $p<0.05$), with no significant difference on MiniCheck or SummaC. Both approaches have similar abstractiveness, with NS achieving scores marginally closer to the human-written abstracts on Trigram novelty. Comparing FS with SH further underscores the necessity of fine-grained local context: SH significantly degrades AlignScore (-0.126), MiniCheck (-0.099), and SummaC (-0.141), all $p < 0.001$, indicating that broad semantic boundaries provided by section headers fail to provide sufficient contextual anchoring for faithful generation.


\begin{table*}[t]
\centering
\small
\setlength{\tabcolsep}{5pt}
\begin{tabular}{lcc|cc|cc|c}
\hline
 & \multicolumn{2}{c}{\textbf{Abstractiveness}} & \multicolumn{2}{c}{\textbf{Factuality}} & \multicolumn{2}{c}{\textbf{Semantic Alignment}} & \\
 \cline{2-3} \cline{4-5} \cline{6-7} \cline{8-8}
\textbf{Config} & \textbf{Tri-g} & \textbf{Dens.} & \textbf{AS} & \textbf{MC} & \textbf{BS} & \textbf{FOCUS$\downarrow$} & \textbf{U-R}\\
\hline
BC & 0.605 (-0.058) & \textbf{4.690 (-1.960)} & \textbf{0.762} & \textbf{0.890} & 0.617 & \textbf{0.828} & \textbf{0.266}\\
\hline
DI & 0.725 (+0.063) & 3.131 (-3.519) & 0.642 & 0.806 & \textbf{0.638} & 0.885 & 0.247\\
DI+Top-5 & 0.725 (+0.062) & 3.138 (-3.512) & 0.644 & 0.807 & 0.637 & 0.933 & 0.245\\
DI+Top-10 & \textbf{0.677 (+0.014)} & 2.938 (-3.712) & 0.644 & 0.811 & 0.595 & 0.915 & 0.231\\
\hline
\end{tabular}
\caption{Ablation on UMLS entity guidance variants. Top-$n$ denotes the number of top-ranked UMLS entities injected into the DI prompt. (Abbreviations as in Table~\ref{tab:eval_primary} and Table~\ref{tab:eval_secondary}) }
\label{tab:eval_umls}
\end{table*}

\begin{table*}[t]
\centering
\small
\setlength{\tabcolsep}{5pt}
\begin{tabular}{lcc|cc|cc|c}
\hline
 & \multicolumn{2}{c}{\textbf{Abstractiveness}} & \multicolumn{2}{c}{\textbf{Factuality}} & \multicolumn{2}{c}{\textbf{Semantic Alignment}} & \\
 \cline{2-3} \cline{4-5} \cline{6-7} \cline{8-8}
\textbf{Config} & \textbf{Tri-g} & \textbf{Dens.} & \textbf{AS} & \textbf{MC} & \textbf{BS} & \textbf{FOCUS$\downarrow$} & \textbf{U-R}\\
\hline
BC & \textbf{0.605 (-0.058)} & \textbf{4.690 (-1.960)} & \textbf{0.762} & \textbf{0.890} & 0.617 & 0.828 & \textbf{0.266}\\
\hline
SI & 0.736 (+0.073) & 3.049 (-3.601) & 0.630 & 0.805 & \textbf{0.636} & 0.836 & 0.251\\
SI + CoT & 0.749 (+0.086) & 2.940 (-3.710) & 0.596 & 0.792 & 0.631 & \textbf{0.794} & 0.249\\
\hline
\end{tabular}
\caption{Ablation on Chain-of-Thought guidance for the SI prompt.  
 (Abbreviations as in Table~\ref{tab:eval_primary} and Table~\ref{tab:eval_secondary} )}
\label{tab:eval_cot}
\end{table*}

\subsection{RQ3: Impact of Prompting Strategy and Entity Guidance}
Next, we examine the effect of increasing prompt complexity.

\paragraph{Prompting Strategy:} While DI and SI yield marginal but significant improvements in BERTScore ($p<0.001$), both suffer from degradation in AlignScore, MiniCheck, SummaC, and UMLS Recall compared to BC (all $p<0.001$; Table~\ref{tab:eval_prompts}). This suggests that dense, multi-faceted instructions might have introduced a distraction effect, where the model struggles to simultaneously satisfy formatting constraints and maintain source-grounded factual alignment, motivating the need for explicit reasoning pathways and external grounding.

\paragraph{Entity Guidance:} To assess whether external grounding can recover the factual consistency degradation observed in DI and SI, we ablate two entity guidance approaches: TR-UMLS integrated into DI, and CoT integrated into SI, with BC (no entity guidance) serving as the reference baseline.

As shown in Table~\ref{tab:eval_umls}, applying TR-UMLS to 
DI does not improve the overall performance. Top-5 and top-10 variants show no significant change in AlignScore relative to the base DI prompt, and both remain significantly below the BC baseline ($p<0.001$). DS-Focus score also increases under top-5 entities ($p<0.01$), suggesting that explicit entity conditioning might cause the model to over-prioritize the provided terms, exacerbating the distraction effect rather than serving as an effective grounding mechanism. 

Table \ref{tab:eval_cot} shows that integrating CoT into the SI prompt improves semantic alignment with the original abstract compared to BC (BERTScore +0.015, $p < 0.001$) while yielding overall higher abstractiveness and lower factuality (AlignScore -0.166, MiniCheck -0.098, SummaC -0.158, all $p<0.001$). This shows that CoT encourages more abstractive and novel phrasing, at the expense of reducing the factual overlap with the source.

\section{Discussion}
Our results show that the proposed training-free approach generates abstracts that more closely resemble human-written abstracts in terms of abstractiveness and factuality. Notably, models fine-tuned on PMC-MAD do not match this performance on these dimensions. However, these fine-tuned models achieve slightly better semantic alignment, lexical/semantic overlap, and compression than DPR-BAG. Since the baseline models do not explicitly model rhetorical structure, we attribute the stronger performance of DPR-BAG on abstractiveness and factuality to its structure-aware design. Specifically, the instruction-tuned LLM backbone provides a baseline tendency toward natural, paraphrased output over verbatim copying. This tendency is amplified by the decompose-then-refine design: partitioning full-text articles into facet-specific sub-documents allows each summarization step to operate over shorter, topically coherent contexts, reducing verbatim copying under long-context pressure. The subsequent refinement stage then re-synthesizes the concatenated facet summaries, further encouraging paraphrasing over fragment assembly.

The distraction effect observed under DI and SI prompts indicates that instruction complexity can actively harm factual grounding in small LLMs. Similarly, entity guidance via UMLS only seemed to increase instruction complexity, yielding little positive effect. The inconsistent gains of SI+CoT further suggest that publication-type distribution can influence generation behavior. This motivates publication-type-aware prompting as a future direction. Additional BC-prompt ablations corroborate these 
findings for TR-UMLS (Appendix~\ref{appendix:bc_entity_guidance}). 

To evaluate the generalizability of DPR-BAG, we applied it to the PubMed Summarization dataset~\cite{pubmed_dataset}. Unlike PMC-MAD, this dataset is not stratified to reflect the publication type distribution of abstract-less PubMed articles. Overall, the trends observed on PMC-MAD, particularly with respect to semantic alignment, abstractiveness, and factuality, largely persist on this dataset. Some differences are also observed (lower MiniCheck scores and better performance of SI+CoT relative to BC). Despite these variations, the results indicate that DPR-BAG maintains robust performance across datasets, supporting its generalizability (Appendix~\ref{appendixx:pubmedSum}).

To investigate whether larger model sizes can improve generation quality, we also evaluated \texttt{Qwen2.5:7B} and \texttt{Qwen2.5:14B} under the BC prompt. These models achieve improvements in semantic alignment and compression rates; however, they also yield excessive abstractiveness and lower factuality, overall yielding little advantage over the base 3B model (Appendix~\ref{appendix:scale-up}).

\subsection{Limitations}
Several limitations apply. First, our evaluation relies primarily on automated metrics, which vary in their robustness. Human evaluation is needed for complementary validation. Second, the BOMRC schema may be suboptimal for articles with non-standard discourse structure. More adaptive decomposition strategies remain an open direction. Finally, DPR-BAG's summaries exhibit substantially higher compression than human-written abstracts, indicating considerably terser outputs. This excessive compression risks omitting auxiliary but informative content such as background, secondary findings, or caveats. Calibrating facet-level length targets is a natural direction for future work.

\section{Related Work}

For the BAG task, \citet{chachra2016} integrated domain-specific classifiers and entailment graphs for extractive sentence selection, inheriting the disjointed flow typical of pure extractive approaches. \citet{cbag} proposed CBAG, which generates abstracts conditioned on author-provided MeSH keywords rather than full-text articles, leaving long-context challenges unaddressed.

To generate abstractive summaries from full-text articles, DANCER~\cite{dancer} and IDCUOT~\cite{idcuot} pioneered the strategy of breaking long documents into manageable sections to bypass context window limitations. However, these supervised approaches rely on heuristic alignment algorithms to map abstract sentences back to source sections and generate summaries for each section in isolation, making them prone to error propagation and fragmented coherence across sections.

GenCompareSum~\cite{gencomparesum} leverages similar divide-and conquer principles but remains extractive-heavy: abstractive fragments serve only as anchors for sentence selection, resulting in discontinuous summaries that lack the narrative transitions typical of human-written abstracts.

\section{Conclusion}
We presented DPR-BAG, a training-free, rhetorical structure-aware, divide-and-conquer framework for biomedical abstract generation. The method decomposes full-text articles into semantic facets and applies parallel LLM-based summarization followed by refinement. Across both the PMC-MAD and PubMed Summarization datasets, DPR-BAG produces abstracts that most closely match human-written abstracts in terms of abstractiveness and factual consistency, without task-specific training and within standard hardware constraints. The framework can be integrated as a preprocessing component into pipelines that require biomedical abstracts, thereby improving downstream performance.

\section*{Acknowledgement}
This work was supported by the National Library of Medicine of the National Institutes of Health under the award number R01LM14292. The content is solely the responsibility of the authors and does not necessarily represent the official views of the National Institutes of Health. The funder had no role in considering the study design or in the collection, analysis, interpretation of data, writing of the report, or decision to submit the article for publication.

\bibliography{custom}

@article{magnet2006letters,
  title={Letters to the editor: Still vigorous after all these years?: A presentation of the discursive and linguistic features of the genre},
  author={Magnet, Anne and Carnet, Didier},
  journal={English for Specific Purposes},
  volume={25},
  number={2},
  pages={173--199},
  year={2006},
  publisher={Elsevier}
}

@article{ioannidis2025house,
  title={In-house editorials and journalistic pieces comprise a massive corpus in the scientific literature that can be improved},
  author={Ioannidis, John PA and Schippers, Micha{\'e}la C},
  journal={European Journal of Clinical Investigation},
  volume={55},
  number={8},
  pages={e70061},
  year={2025},
  publisher={Wiley Online Library}
}

@article{gurulingappa2012development,
  title={Development of a benchmark corpus to support the automatic extraction of drug-related adverse effects from medical case reports},
  author={Gurulingappa, Harsha and Rajput, Abdul Mateen and Roberts, Angus and Fluck, Juliane and Hofmann-Apitius, Martin and Toldo, Luca},
  journal={Journal of biomedical informatics},
  volume={45},
  number={5},
  pages={885--892},
  year={2012},
  publisher={Elsevier}
}

@inproceedings{
splittingmakesstronger,
title={When Splitting Makes Stronger: A Theoretical and Empirical Analysis of Divide-and-Conquer Prompting in {LLM}s},
author={Yizhou Zhang and Defu Cao and Lun Du and Qiang Fu and Yan Liu},
booktitle={Second Conference on Language Modeling},
year={2025},
url={https://openreview.net/forum?id=rAR7iPI8Kh}
}

@article{scholarSum,
author = {Wang, Tairan and Chen, Xiuying and Zhu, Qingqing and Guo, Taicheng and Gao, Shen and Lu, Zhiyong and Gao, Xin and Zhang, Xiangliang},
title = {New Paradigm for Evaluating Scholar Summaries: A Facet-aware Metric and a Meta-evaluation Benchmark},
year = {2025},
issue_date = {July 2025},
publisher = {Association for Computing Machinery},
address = {New York, NY, USA},
volume = {43},
number = {4},
issn = {1046-8188},
url = {https://doi.org/10.1145/3733597},
doi = {10.1145/3733597},
journal = {ACM Trans. Inf. Syst.},
month = jun,
articleno = {100},
numpages = {25}
}

@ARTICLE{abstractiveSum,
  author={Chernyshev, Daniil and Dobrov, Boris},
  journal={IEEE Access}, 
  title={Investigating the Pre-Training Bias in Low-Resource Abstractive Summarization}, 
  year={2024},
  volume={12},
  number={},
  pages={47219-47230},
  doi={10.1109/ACCESS.2024.3379139}}

@ARTICLE{dancer,
  author={Gidiotis, Alexios and Tsoumakas, Grigorios},
  journal={IEEE/ACM Transactions on Audio, Speech, and Language Processing}, 
  title={A Divide-and-Conquer Approach to the Summarization of Long Documents}, 
  year={2020},
  volume={28},
  number={},
  pages={3029-3040},
  doi={10.1109/TASLP.2020.3037401}}

@INPROCEEDINGS{idcuot,
  author={Shen, Xin and Lam, Wai},
  booktitle={2022 4th International Conference on Natural Language Processing (ICNLP)}, 
  title={Improved Divide-and-Conquer Approach to Abstractive Summarization of Scientific Papers}, 
  year={2022},
  volume={},
  number={},
  pages={395-398},
  doi={10.1109/ICNLP55136.2022.00073}}

@inproceedings{secnormposter,
  author = {Lin, Sylvey and Menke, Joseph and Holt, Arthur and Kilicoglu, Halil and Smalheiser, Neil},
  title = {Section Header Normalization in Biomedical Articles using Transformers},
  booktitle = {AMIA Annual Symposium Proceedings},
  year = {2025},
  note = {Poster P116},
  url = {https://amia.secure-platform.com/symposium/gallery/rounds/82021/details/19558}
}

@inproceedings{LLM-SSC,
    title = "Multi-label Sequential Sentence Classification via Large Language Model",
    author = "Lan, Mengfei  and
      Zheng, Lecheng  and
      Ming, Shufan  and
      Kilicoglu, Halil",
    editor = "Al-Onaizan, Yaser  and
      Bansal, Mohit  and
      Chen, Yun-Nung",
    booktitle = "Findings of the Association for Computational Linguistics: EMNLP 2024",
    month = nov,
    year = "2024",
    address = "Miami, Florida, USA",
    publisher = "Association for Computational Linguistics",
    url = "https://aclanthology.org/2024.findings-emnlp.944/",
    doi = "10.18653/v1/2024.findings-emnlp.944",
    pages = "16086--16104"
}

@inproceedings{discoscore,
    title = "{D}isco{S}core: Evaluating Text Generation with {BERT} and Discourse Coherence",
    author = "Zhao, Wei  and
      Strube, Michael  and
      Eger, Steffen",
    editor = "Vlachos, Andreas  and
      Augenstein, Isabelle",
    booktitle = "Proceedings of the 17th Conference of the European Chapter of the Association for Computational Linguistics",
    month = may,
    year = "2023",
    address = "Dubrovnik, Croatia",
    publisher = "Association for Computational Linguistics",
    url = "https://aclanthology.org/2023.eacl-main.278/",
    doi = "10.18653/v1/2023.eacl-main.278",
    pages = "3865--3883"
}

@inproceedings{newsroom,
    title = "{N}ewsroom: A Dataset of 1.3 Million Summaries with Diverse Extractive Strategies",
    author = "Grusky, Max  and
      Naaman, Mor  and
      Artzi, Yoav",
    editor = "Walker, Marilyn  and
      Ji, Heng  and
      Stent, Amanda",
    booktitle = "Proceedings of the 2018 Conference of the North {A}merican Chapter of the Association for Computational Linguistics: Human Language Technologies, Volume 1 (Long Papers)",
    month = jun,
    year = "2018",
    address = "New Orleans, Louisiana",
    publisher = "Association for Computational Linguistics",
    url = "https://aclanthology.org/N18-1065/",
    doi = "10.18653/v1/N18-1065",
    pages = "708--719"
}

@inproceedings{novelty,
    title = "Get To The Point: Summarization with Pointer-Generator Networks",
    author = "See, Abigail  and
      Liu, Peter J.  and
      Manning, Christopher D.",
    editor = "Barzilay, Regina  and
      Kan, Min-Yen",
    booktitle = "Proceedings of the 55th Annual Meeting of the Association for Computational Linguistics (Volume 1: Long Papers)",
    month = jul,
    year = "2017",
    address = "Vancouver, Canada",
    publisher = "Association for Computational Linguistics",
    url = "https://aclanthology.org/P17-1099/",
    doi = "10.18653/v1/P17-1099",
    pages = "1073--1083"
}

@misc{beltagy2020longformer,
      title={Longformer: The Long-Document Transformer}, 
      author={Iz Beltagy and Matthew E. Peters and Arman Cohan},
      year={2020},
      eprint={2004.05150},
      archivePrefix={arXiv},
      primaryClass={cs.CL},
      url={https://arxiv.org/abs/2004.05150}, 
}

@inproceedings{pubmed_dataset,
  title = "A Discourse-Aware Attention Model for Abstractive Summarization of Long Documents",
  author = "Cohan, Arman  and
    Dernoncourt, Franck  and
    Kim, Doo Soon  and
    Bui, Trung  and
    Kim, Seokhwan  and
    Chang, Walter  and
    Goharian, Nazli",
  booktitle = "Proceedings of the 2018 Conference of the North {A}merican Chapter of the Association for Computational Linguistics: Human Language Technologies, Volume 2 (Short Papers)",
  month = jun,
  year = "2018",
  address = "New Orleans, Louisiana",
  publisher = "Association for Computational Linguistics",
  url = "https://aclanthology.org/N18-2097",
  doi = "10.18653/v1/N18-2097",
  pages = "615--621",
}

@article{ir1, title={Improving Biomedical Information Retrieval with Neural Retrievers}, volume={36}, url={https://ojs.aaai.org/index.php/AAAI/article/view/21352}, DOI={10.1609/aaai.v36i10.21352}, abstractNote={Information retrieval (IR) is essential in search engines and dialogue systems as well as natural language processing tasks such as open-domain question answering. IR serve an important function in the biomedical domain, where content and sources of scientific knowledge may evolve rapidly. Although neural retrievers have surpassed traditional IR approaches such as TF-IDF and BM25 in standard open-domain question answering tasks, they are still found lacking in the biomedical domain. In this paper, we seek to improve information retrieval (IR) using neural retrievers (NR) in the biomedical domain, and achieve this goal using a three-pronged approach. First, to tackle the relative lack of data in the biomedical domain, we propose a template-based question generation method that can be leveraged to train neural retriever models. Second, we develop two novel pre-training tasks that are closely aligned to the downstream task of information retrieval. Third, we introduce the ``Poly-DPR’’ model which encodes each context into multiple context vectors. Extensive experiments and analysis on the BioASQ challenge suggest that our proposed method leads to large gains over existing neural approaches and beats BM25 in the small-corpus setting. We show that BM25 and our method can complement each other, and a simple hybrid model leads to further gains in the large corpus setting.}, number={10}, journal={Proceedings of the AAAI Conference on Artificial Intelligence}, author={Luo, Man and Mitra, Arindam and Gokhale, Tejas and Baral, Chitta}, year={2022}, month={Jun.}, pages={11038-11046} }

@inproceedings{ir2,
author = {Ueda, Alberto and Santos, Rodrygo L. T. and Macdonald, Craig and Ounis, Iadh},
title = {Structured Fine-Tuning of Contextual Embeddings for Effective Biomedical Retrieval},
year = {2021},
isbn = {9781450380379},
publisher = {Association for Computing Machinery},
address = {New York, NY, USA},
url = {https://doi.org/10.1145/3404835.3463075},
doi = {10.1145/3404835.3463075},
booktitle = {Proceedings of the 44th International ACM SIGIR Conference on Research and Development in Information Retrieval},
pages = {2031–2035},
numpages = {5},
location = {Virtual Event, Canada},
series = {SIGIR '21}
}

@article{biocuration1,
    author = {Wiegers, Thomas C and Davis, Allan Peter and Wiegers, Jolene and Sciaky, Daniela and Barkalow, Fern and Wyatt, Brent and Strong, Melissa and McMorran, Roy and Abrar, Sakib and Mattingly, Carolyn J},
    title = {Integrating {AI}-powered text mining from {P}ub{T}ator into the manual curation workflow at the {C}omparative {T}oxicogenomics {D}atabase},
    journal = {Database},
    volume = {2025},
    pages = {baaf013},
    year = {2025},
    month = {02},
    issn = {1758-0463},
    doi = {10.1093/database/baaf013},
    url = {https://doi.org/10.1093/database/baaf013},
    eprint = {https://academic.oup.com/database/article-pdf/doi/10.1093/database/baaf013/62047329/baaf013.pdf},
}

@inproceedings{pubmedQA,
  title={{P}ub{M}ed{QA}: A Dataset for Biomedical Research Question Answering},
  author={Jin, Qiao and Dhingra, Bhuwan and Liu, Zhengping and Cohen, William and Lu, Xinghua},
  booktitle={Proceedings of the 2019 Conference on Empirical Methods in Natural Language Processing and the 9th International Joint Conference on Natural Language Processing (EMNLP-IJCNLP)},
  pages={2567--2577},
  year={2019}
}

@Article{a_vs_e,
AUTHOR = {Giarelis, Nikolaos and Mastrokostas, Charalampos and Karacapilidis, Nikos},
TITLE = {Abstractive vs. Extractive Summarization: An Experimental Review},
JOURNAL = {Applied Sciences},
VOLUME = {13},
YEAR = {2023},
NUMBER = {13},
ARTICLE-NUMBER = {7620},
URL = {https://www.mdpi.com/2076-3417/13/13/7620},
ISSN = {2076-3417},
DOI = {10.3390/app13137620}
}

@inproceedings{gencomparesum,
    title = "{G}en{C}ompare{S}um: a hybrid unsupervised summarization method using salience",
    author = "Bishop, Jennifer  and
      Xie, Qianqian  and
      Ananiadou, Sophia",
    editor = "Demner-Fushman, Dina  and
      Cohen, Kevin Bretonnel  and
      Ananiadou, Sophia  and
      Tsujii, Junichi",
    booktitle = "Proceedings of the 21st Workshop on Biomedical Language Processing",
    month = may,
    year = "2022",
    address = "Dublin, Ireland",
    publisher = "Association for Computational Linguistics",
    url = "https://aclanthology.org/2022.bionlp-1.22/",
    doi = "10.18653/v1/2022.bionlp-1.22",
    pages = "220--240"
}

@inproceedings{longt5,
    title = "{L}ong{T}5: {E}fficient Text-To-Text Transformer for Long Sequences",
    author = "Guo, Mandy  and
      Ainslie, Joshua  and
      Uthus, David  and
      Onta{\~n}{\'o}n, Santiago  and
      Ni, Jianmo  and
      Sung, Yun-Hsuan  and
      Yang, Yinfei",
    editor = "Carpuat, Marine  and
      de Marneffe, Marie-Catherine  and
      Meza Ruiz, Ivan Vladimir",
    booktitle = "Findings of the Association for Computational Linguistics: NAACL 2022",
    month = jul,
    year = "2022",
    address = "Seattle, United States",
    publisher = "Association for Computational Linguistics",
    url = "https://aclanthology.org/2022.findings-naacl.55/",
    doi = "10.18653/v1/2022.findings-naacl.55",
    pages = "724--736"
}

@article{case_report,
title = {Adverse drug event detection and extraction from open data: A deep learning approach},
journal = {Information Processing \& Management},
volume = {57},
number = {1},
pages = {102131},
year = {2020},
issn = {0306-4573},
doi = {https://doi.org/10.1016/j.ipm.2019.102131},
url = {https://www.sciencedirect.com/science/article/pii/S0306457319308623},
author = {Brandon Fan and Weiguo Fan and Carly Smith and Harold ``Skip'' Garner}
}

@article{pubmedBERT,
author = {Gu, Yu and Tinn, Robert and Cheng, Hao and Lucas, Michael and Usuyama, Naoto and Liu, Xiaodong and Naumann, Tristan and Gao, Jianfeng and Poon, Hoifung},
title = {Domain-Specific Language Model Pretraining for Biomedical Natural Language Processing},
year = {2021},
issue_date = {January 2022},
publisher = {Association for Computing Machinery},
address = {New York, NY, USA},
volume = {3},
number = {1},
url = {https://doi.org/10.1145/3458754},
doi = {10.1145/3458754},
journal = {ACM Trans. Comput. Healthcare},
month = oct,
articleno = {2},
numpages = {23}
}

@article{letters,
  author  = {Nuzzo, James L.},
  title   = {Letters to the editor in exercise science and physical therapy journals: an examination of content and ``authorship inflation''},
  journal = {Scientometrics},
  year    = {2021},
  volume  = {126},
  number  = {8},
  pages   = {6917--6936},
  doi     = {10.1007/s11192-021-04068-w},
  url     = {https://doi.org/10.1007/s11192-021-04068-w},
  issn    = {1588-2861}
}

@article{Waaijer2011,
  author   = {Waaijer, Cathelijn J. F. and van Bochove, Cornelis A. and van Eck, Nees Jan},
  title    = {On the map: Nature and Science editorials},
  journal  = {Scientometrics},
  year     = {2011},
  volume   = {86},
  number   = {1},
  pages    = {99--112},
  doi      = {10.1007/s11192-010-0205-9},
  url      = {https://doi.org/10.1007/s11192-010-0205-9},
  issn     = {1588-2861}
}

@inproceedings{chachra2016,
    title = "A Hybrid Approach to Generation of Missing Abstracts in Biomedical Literature",
    author = "Chachra, Suchet  and
      Ben Abacha, Asma  and
      Shooshan, Sonya  and
      Rodriguez, Laritza  and
      Demner-Fushman, Dina",
    editor = "Matsumoto, Yuji  and
      Prasad, Rashmi",
    booktitle = "Proceedings of {COLING} 2016, the 26th International Conference on Computational Linguistics: Technical Papers",
    month = dec,
    year = "2016",
    address = "Osaka, Japan",
    publisher = "The COLING 2016 Organizing Committee",
    url = "https://aclanthology.org/C16-1104/",
    pages = "1093--1100"
}

@article{multitagger,
  title={Publication Type Tagging using Transformer Models and Multi-Label Classification},
  author={Menke, Joe D and Kilicoglu, Halil and Smalheiser, Neil R},
  journal={AMIA Annual Symposium Proceedings},
  volume={2024},
  pages={818--827},
  year={2024},
  publisher={American Medical Informatics Association},
  pmid={40417522},
  pmcid={PMC12099436}
}

@misc{formattingimpact,
      title={Does Prompt Formatting Have Any Impact on {LLM} Performance?}, 
      author={Jia He and Mukund Rungta and David Koleczek and Arshdeep Sekhon and Franklin X Wang and Sadid Hasan},
      year={2024},
      eprint={2411.10541},
      archivePrefix={arXiv},
      primaryClass={cs.CL},
      url={https://arxiv.org/abs/2411.10541}, 
}

@article{summaC,
    author = {Laban, Philippe and Schnabel, Tobias and Bennett, Paul N. and Hearst, Marti A.},
    title = {{S}umma{C}: Re-Visiting {NLI}-based Models for Inconsistency Detection in Summarization},
    journal = {Transactions of the Association for Computational Linguistics},
    volume = {10},
    pages = {163-177},
    year = {2022},
    month = {02},
    issn = {2307-387X},
    doi = {10.1162/tacl_a_00453},
    url = {https://doi.org/10.1162/tacl_a_00453},
    eprint = {https://direct.mit.edu/tacl/article-pdf/doi/10.1162/tacl_a_00453/1987014/tacl_a_00453.pdf},
}

@inproceedings{alignscore,
    title = "{A}lign{S}core: Evaluating Factual Consistency with A Unified Alignment Function",
    author = "Zha, Yuheng  and
      Yang, Yichi  and
      Li, Ruichen  and
      Hu, Zhiting",
    editor = "Rogers, Anna  and
      Boyd-Graber, Jordan  and
      Okazaki, Naoaki",
    booktitle = "Proceedings of the 61st Annual Meeting of the Association for Computational Linguistics (Volume 1: Long Papers)",
    month = jul,
    year = "2023",
    address = "Toronto, Canada",
    publisher = "Association for Computational Linguistics",
    url = "https://aclanthology.org/2023.acl-long.634/",
    doi = "10.18653/v1/2023.acl-long.634",
    pages = "11328--11348"
}

@inproceedings{minicheck,
    title = "{M}ini{C}heck: Efficient Fact-Checking of {LLM}s on Grounding Documents",
    author = "Tang, Liyan  and
      Laban, Philippe  and
      Durrett, Greg",
    editor = "Al-Onaizan, Yaser  and
      Bansal, Mohit  and
      Chen, Yun-Nung",
    booktitle = "Proceedings of the 2024 Conference on Empirical Methods in Natural Language Processing",
    month = nov,
    year = "2024",
    address = "Miami, Florida, USA",
    publisher = "Association for Computational Linguistics",
    url = "https://aclanthology.org/2024.emnlp-main.499/",
    doi = "10.18653/v1/2024.emnlp-main.499",
    pages = "8818--8847"
}

@article{cbag,
  title={{CBAG}: Conditional biomedical abstract generation},
  author={Sybrandt, Justin and Safro, Ilya},
  journal={PLoS One},
  volume={16},
  number={7},
  pages={e0253905},
  year={2021},
  publisher={Public Library of Science San Francisco, CA USA},
  doi={10.1371/journal.pone.0253905},
  url={https://doi.org/10.1371/journal.pone.0253905}
}

@inproceedings{pubmed200krct,
    title = "{P}ub{M}ed 200k {RCT}: a Dataset for Sequential Sentence Classification in Medical Abstracts",
    author = "Dernoncourt, Franck  and
      Lee, Ji Young",
    editor = "Kondrak, Greg  and
      Watanabe, Taro",
    booktitle = "Proceedings of the Eighth International Joint Conference on Natural Language Processing (Volume 2: Short Papers)",
    month = nov,
    year = "2017",
    address = "Taipei, Taiwan",
    publisher = "Asian Federation of Natural Language Processing",
    url = "https://aclanthology.org/I17-2052/",
    pages = "308--313"
}

@inproceedings{scispacy,
    title = "{S}cispa{C}y: {F}ast and {R}obust {M}odels for {B}iomedical {N}atural {L}anguage {P}rocessing",
    author = "Neumann, Mark  and
      King, Daniel  and
      Beltagy, Iz  and
      Ammar, Waleed",
    booktitle = "Proceedings of the 18th BioNLP Workshop and Shared Task",
    month = aug,
    year = "2019",
    address = "Florence, Italy",
    publisher = "Association for Computational Linguistics",
    url = "https://www.aclweb.org/anthology/W19-5034",
    doi = "10.18653/v1/W19-5034",
    pages = "319--327",
    eprint = {arXiv:1902.07669},
}

@article{umls,
    title   = {The {Unified Medical Language System} ({UMLS}): integrating biomedical terminology},
    author  = {Bodenreider, Olivier},
    journal = {Nucleic Acids Research},
    volume  = {32},
    number  = {Database issue},
    pages   = {D267--D270},
    year    = {2004},
    month   = {1},
    doi     = {10.1093/nar/gkh061},
    pmid    = {14681409},
    pmcid   = {PMC308795}
}

@inproceedings{rouge,
    title = "{ROUGE}: A Package for Automatic Evaluation of Summaries",
    author = "Lin, Chin-Yew",
    booktitle = "Text Summarization Branches Out",
    month = jul,
    year = "2004",
    address = "Barcelona, Spain",
    publisher = "Association for Computational Linguistics",
    url = "https://aclanthology.org/W04-1013/",
    pages = "74--81"
}

@inproceedings{
bertscore,
title={{BERTS}core: Evaluating Text Generation with {BERT}},
author={Tianyi Zhang and Varsha Kishore and Felix Wu and Kilian Q. Weinberger and Yoav Artzi},
booktitle={International Conference on Learning Representations},
year={2020},
url={https://openreview.net/forum?id=SkeHuCVFDr}
}

@inproceedings{cohanpretrained,
    title = "Pretrained Language Models for Sequential Sentence Classification",
    author = "Cohan, Arman  and
      Beltagy, Iz  and
      King, Daniel  and
      Dalvi, Bhavana  and
      Weld, Dan",
    editor = "Inui, Kentaro  and
      Jiang, Jing  and
      Ng, Vincent  and
      Wan, Xiaojun",
    booktitle = "Proceedings of the 2019 Conference on Empirical Methods in Natural Language Processing and the 9th International Joint Conference on Natural Language Processing (EMNLP-IJCNLP)",
    month = nov,
    year = "2019",
    address = "Hong Kong, China",
    publisher = "Association for Computational Linguistics",
    url = "https://aclanthology.org/D19-1383/",
    doi = "10.18653/v1/D19-1383",
    pages = "3693--3699"
}

\appendix
\section{Summarization Prompt Templates}
\label{appendix:prompts}

All prompting strategies share a \texttt{facet\_guidelines} dictionary that maps each facet label to a facet-specific instruction. We use two versions: a concise version (Table~\ref{tab:guidelines-short}) used in BC and SI prompts, and a detailed version adapted from JMIR author guidelines (see footnote \ref{footnote:jmir}) (Table~\ref{tab:guidelines-long} used in DI prompts). In all prompt templates, \texttt{<facet\_text>} denotes the input facet text, \texttt{<facet\_type>} denotes the rhetorical facet label, and \texttt{<facet\_guide>} denotes the corresponding facet-specific instruction.

\begin{table*}
\centering
\small
\begin{tabular}{ll}
\toprule
\textbf{Facet} & \textbf{Guideline} \\
\midrule
Background & Focus on research gap and motivation. \\
Objective & State the primary aim or hypothesis. \\
Methods & Detail study design and procedures. \\
Results & Prioritize key findings and data. \\
Conclusions & Summarize implications and take-home messages. \\
Others & Focus on the main point of the paragraph. \\
Intro & Introduce the main topic and context. \\
Main Idea & Focus on the central concept or hypothesis. \\
Results \& Conclusions & Focus on key findings and their implications. \\
\bottomrule
\end{tabular}
\caption{Concise facet-specific guidelines (used in BC and SI prompts).}
\label{tab:guidelines-short}
\end{table*}

\begin{table*}
\centering
\small

\begin{tabular}{lp{8cm}}
\toprule
\textbf{Facet} & \textbf{Guideline} \\
\midrule
Background & Briefly describe the context and significance of the research. \\
Objective & State the specific aim(s) of the study in a complete sentence. \\
Methods & Outline the research design, study sample, data collection, and analysis procedures. \\
Results & Present key findings, including relevant statistics (sample sizes, response rates, P values, confidence intervals). Be specific. \\
Conclusions & Summarize the main findings and their implications. \\
Others & Synthesize the core biomedical information, focusing on the primary argument, concept, or supplementary context presented. \\
Intro & Describe the research context and significance, and clearly state the specific aims or hypotheses. \\
Main Idea & Outline the research design and procedures, while capturing any supplementary methodological context or core concepts. \\
Results \& Conclusions & Present key findings with relevant statistics, and summarize their broader implications and take-home messages. \\
\bottomrule
\end{tabular}
\caption{Detailed facet-specific guidelines (used for DI prompts), adapted from JMIR author guidelines.}
\label{tab:guidelines-long}
\end{table*}

\subsection{Basic Concise Prompt (BC)}
\label{appendix:prompt-bs}

\begin{tcolorbox}[colback=gray!8, colframe=gray!40, title=System Message, fonttitle=\bfseries]
You are a biomedical summarization assistant. 1. Use a formal, objective, scientific tone. 2. Never use meta-phrases like 'the authors state' or 'this section describes'.
Respond ONLY with a JSON object: 
            \{"summary": "...", "reasoning": "..."\}. 
            No markdown, no talk.
\end{tcolorbox}

\begin{tcolorbox}[colback=blue!5, colframe=blue!30, title=User Message, fonttitle=\bfseries]
Summarize this \texttt{<facet\_type>} section\\
Specific focus: \texttt{<facet\_guide>}\\\\
Paragraph text:\\
\texttt{<facet\_text>}
\end{tcolorbox}

\subsection{Detailed Instruction Prompt (DI)}
\label{appendix:prompt-di}

\begin{tcolorbox}[colback=gray!8, colframe=gray!40, title=System Message, fonttitle=\bfseries]
You are a biomedical synthesis assistant. 1. Use a formal, objective, scientific tone. \\ 2. Never use meta-phrases like `the authors state' or `this section describes'. \\ Define your output strictly as:\\ -`summary': The synthesized biomedical text. \\-`reasoning': A brief explanation. \\Format: \{"summary": "...", "reasoning": "..."\}. No markdown, no talk.
\end{tcolorbox}

\begin{tcolorbox}[colback=blue!5, colframe=blue!30, title=User Message, fonttitle=\bfseries]
Synthesize the critical information from this \texttt{<facet\_type>} section provided in the `Paragraph text'\\
\texttt{<facet\_guide>}\\\\
Paragraph text:\\
\texttt{<facet\_text>}
\end{tcolorbox}

\subsection{Structural Instruction Prompt (SI)}
\label{appendix:prompt-si}

\begin{tcolorbox}[colback=gray!8, colframe=gray!40, title=System Message, fonttitle=\bfseries]
\# ROLE\\
You are an expert Biomedical Summarization Assistant.\\\\
\# STRICT GUIDELINES\\
- **Tone**: Formal, objective, and academic.\\
- **No Meta-Talk**: Do NOT use phrases like `The authors state' or `This section describes'.\\
- **Output Format**: Respond **ONLY** with a valid JSON object. No Markdown blocks, no preamble, no postscript.\\
```json\\
\{"reasoning": "Brief explanation of how your summary fulfills the given instructions...", "summary": "Final summary..."\}
```
\end{tcolorbox}

\begin{tcolorbox}[colback=blue!5, colframe=blue!30, title=User Message, fonttitle=\bfseries]
\#\# TASK: SUMMARY GENERATION\\
\ **Target Focus:** \texttt{<facet\_guide>}\\\\
\ **Instructions:** Generate a professional biomedical summary.\\
---\\
\#\#\# INPUT TEXT (Reference)\\
\texttt{<facet\_text>}
\end{tcolorbox}

\subsection{BC for Naive Splitting (BC-NS)}
\label{appendix:prompt-bs-ns}

When paired with the Naive Splitting baseline, the system message remains unchanged. As no facet label is assigned, the user message uses a generic focus instruction:

\begin{tcolorbox}[colback=blue!5, colframe=blue!30, title=User Message, fonttitle=\bfseries]
Summarize this section.\\
Specific focus: Summarize the main topic.\\\\
Paragraph text:\\
\texttt{<facet\_text>}
\end{tcolorbox}

\subsection{BC with TR-UMLS Entity Guidance (BC+TR-UMLS)}
\label{appendix:prompt-bc-umls}

\begin{tcolorbox}[colback=gray!8, colframe=gray!40, title=System Message, fonttitle=\bfseries]
You are a biomedical summarization assistant. 1. Use a formal, objective, scientific tone. 2. Never use meta-phrases like 'the authors state' or 'this section describes'.
Respond ONLY with a JSON object: 
            \{"summary": "...", "reasoning": "..."\}. 
            No markdown, no talk.
\end{tcolorbox}

\begin{tcolorbox}[colback=blue!5, colframe=blue!30, title=User Message, fonttitle=\bfseries]
Summarize this \texttt{<facet\_type>} section\\
Specific focus: \texttt{<facet\_guide>}\\
Ensure the core meanings of these key biomedical entities are preserved or synthesized accurately: \\\\
Paragraph text:<top\_entities>\\
\texttt{<facet\_text>}
\end{tcolorbox}

\subsection{DI with TR-UMLS Entity Guidance (DI+TR-UMLS)}
\label{appendix:prompt-di-umls}


\begin{tcolorbox}[colback=blue!5, colframe=blue!30, title=User Message, fonttitle=\bfseries]
Synthesize the critical information from this \texttt{<facet\_type>} section provided in the `Paragraph text'\\
\texttt{<facet\_guide>}\\\\
Ensure the core meanings of these key biomedical entities are preserved or synthesized accurately: <top\_entities>\\\\
Paragraph text:\\
\texttt{<facet\_text>}
\end{tcolorbox}

\subsection{SI with Chain-of-Thought (SI+CoT)}
\label{appendix:prompt-si-cot}

\begin{tcolorbox}[colback=gray!8, colframe=gray!40, title=System Message, fonttitle=\bfseries]
\# ROLE\\
You are an expert Biomedical Summarization Assistant.\\\\
--- \\
\# OPERATIONAL FRAMEWORK\\
You must follow this 2-Stage Chain-of-Thought process:\\
1. Stage 1: Element Extraction (entities, parameters, methodologies, statistics).\\
2. Stage 2: Summary Generation (synthesize into scientific narrative).\\\\
--- \\
\# STRICT GUIDELINES\\
- **Tone**: Formal, objective, and academic.\\
- **No Meta-Talk**: Do NOT use phrases like `The authors state' or `This section describes'.\\
- **Output Format**: Respond **ONLY** with a valid JSON object. No Markdown blocks, no preamble, no postscript.\\
```json\\
\{"reasoning": "Stage 1 extraction results...", "summary": "Stage 2 final summary..."\}\\
```
\end{tcolorbox}

\begin{tcolorbox}[colback=blue!5, colframe=blue!30, title=User Message 1 (Stage 1), fonttitle=\bfseries]
\#\# TASK 1: ELEMENT EXTRACTION\\
\ **Section Type:** \texttt{<facet\_type>}\\
\ **Instructions:** Extract key elements from the text below, including:\\
- **Entities:** Diseases, genes, drugs, proteins.\\
- **Parameters:** Sample sizes, dosage, duration.\\
- **Methodology:** Study design, assays, equipment.\\
- **Statistics:** P-values, confidence intervals, effect sizes.\\\\
---\\
\#\#\# INPUT TEXT\\
\texttt{<facet\_text>}
\end{tcolorbox}

\begin{tcolorbox}[colback=blue!5, colframe=blue!30, title=User Message 2 (Stage 2), fonttitle=\bfseries]
\#\# TASK 2: SUMMARY GENERATION\\
\ **Target Focus:** \texttt{<facet\_guide>}\\\\
\ **Instructions:** Using the elements extracted in Task 1, generate a professional biomedical summary.\\
Ensure the summary is dense with information but remains readable and scientifically accurate.\\\\
---\\
\#\#\# INPUT TEXT (Reference)\\
\texttt{<facet\_text>}
\end{tcolorbox}

\section{Validation and Fallback Details}
\label{appendix:fallback}

When the fallback process is triggered, the original six facets are regrouped into three broader categories: \textit{Intro} (Background and Objective), \textit{Main Idea} (Methods and Others), and \textit{Results \& Conclusions} (Results and Conclusions). Background and Objective are merged as both establish the research context and motivation. Results and Conclusions are paired as both convey findings and their implications. The remaining Others facet, which retains paragraphs not assigned to any BOMRC category by the sentence classifier, is grouped with Methods by elimination, as the other four facets form more natural rhetorical pairs. These regrouped facets are then sent back to the parallel summarization module. If the LLM still fails to summarize a specific regrouped facet, the first 300 characters (Lead-300) of that facet are used as the facet summary to capture core information via lead bias without adding granular noise. The fallback mechanism was rarely triggered in practice, suggesting that this pairing has minimal impact on overall generation quality. In a random sample of 300 test articles (6.5\% of the test set), the fallback mechanism was never triggered, suggesting with 95\% confidence that fewer than 1\% of articles require fallback intervention.

\section{Refinement Prompt}
\label{appendix:refined_prompt}

This prompt is used in the final stage to smooth the concatenated facets, ensuring structural coherence and stylistic consistency across the unified abstract. \texttt{<draft\_abstract>} denotes the place holder for the concatenated draft abstract.

\begin{tcolorbox}[colback=gray!8, colframe=gray!40, title=System Message, fonttitle=\bfseries]
You are a biomedical abstract refinement assistant. Refine the abstract based on the abstract draft.\\
    CRITICAL INSTRUCTION:\\
    Respond ONLY with a valid JSON object. \\
    Do NOT use Markdown code blocks (like ```json). \\
    Do NOT provide any conversational text.\\
    \\
    Format:\\
    \{\\
        "abstract": "your abstract text here",\\
        "reasoning": "your reasoning here"\\
    \}
\end{tcolorbox}

\begin{tcolorbox}[colback=blue!5, colframe=blue!30, title=User Message, fonttitle=\bfseries]
abstract draft: \texttt{<draft\_abstract>}
\end{tcolorbox}

\section{Extended Implementation Details and Token Distribution}
\label{appendix:implementation_details}

All fine-tuning and evaluation procedures were conducted on an NVIDIA Tesla V100 GPU (32GB VRAM). While the underlying LED architecture theoretically supports sequences up to 16,384 tokens, processing such lengths on standard hardware is computationally prohibitive, inevitably leading to out-of-memory (OOM) errors even with minimal batch sizes. To fit within this memory budget, the fine-tuned LED-Pubmed used gradient accumulation with an effective batch size of 8, halted at 500 steps based on validation performance.

\begin{figure}
    \centering
    \includegraphics[width=0.5\textwidth]{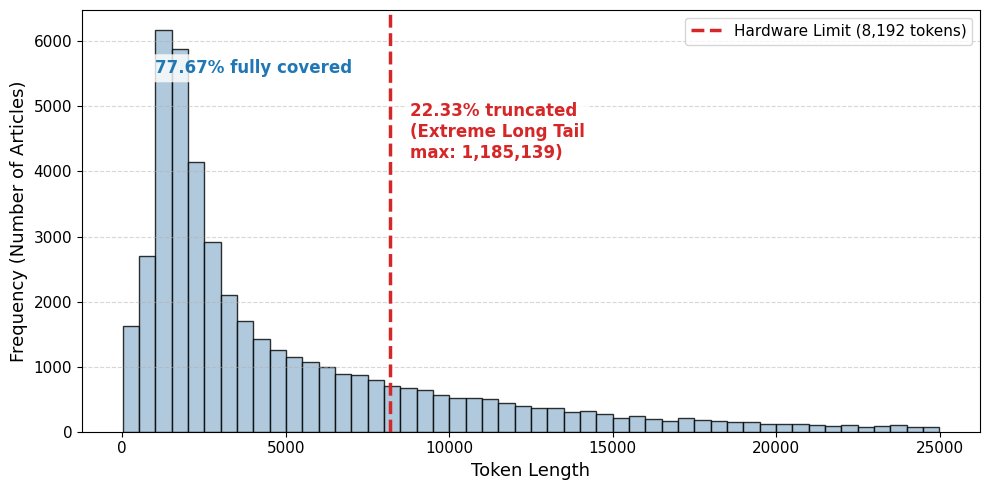} 
    \caption{Distribution of document token lengths in the dataset. The red dashed line denotes the 8,192-token hardware limit for standard baselines.}
    \label{fig:distribution}
\end{figure}

This constraint restricts standard baselines to 8,192-token inputs. As illustrated in Figure \ref{fig:distribution}, our empirical analysis of the 46,309 articles in the dataset reveals a median of 2,959 tokens and an average length of approximately 6,018 tokens. While the 8,192-token capacity successfully accommodates 77.67\% of the dataset, the length distribution exhibits a severe long-tail characteristic. Specifically, 22.33\% of the articles exceed this limit, with the longest document reaching an extreme 1,185,139 tokens. For these extensive studies, standard baselines operating within memory limits are forced to truncate critical information, such as discussion and conclusion sections, which underscores the necessity of the partitioned approach introduced in our DPR-BAG framework.

\section{LLM-SSC}
\label{appendix:LLM-SSC}
LLM-SSC~\cite{LLM-SSC} is evaluated on the BIORC800 dataset, a manually annotated multi-label SSC dataset of biomedical abstracts using the BOMRC schema. Under task-specific fine-tuning, LLM-SSC achieves a micro F1 of 0.907 and macro F1 of 0.912 on BIORC800, outperforming prior SSC baselines~\cite{cohanpretrained}. We adopt LLM-SSC for our document splitting module as its label schema (Background, Objective, Methods, Results, Conclusions, and None) directly aligns with the BOMRC+ facets in DPR-BAG, where the None label corresponds to our Others facet.

\section{PubMedSum Dataset Validation}  
\label{appendixx:pubmedSum}

\begin{figure}
    \centering
    \includegraphics[width=0.5\textwidth]{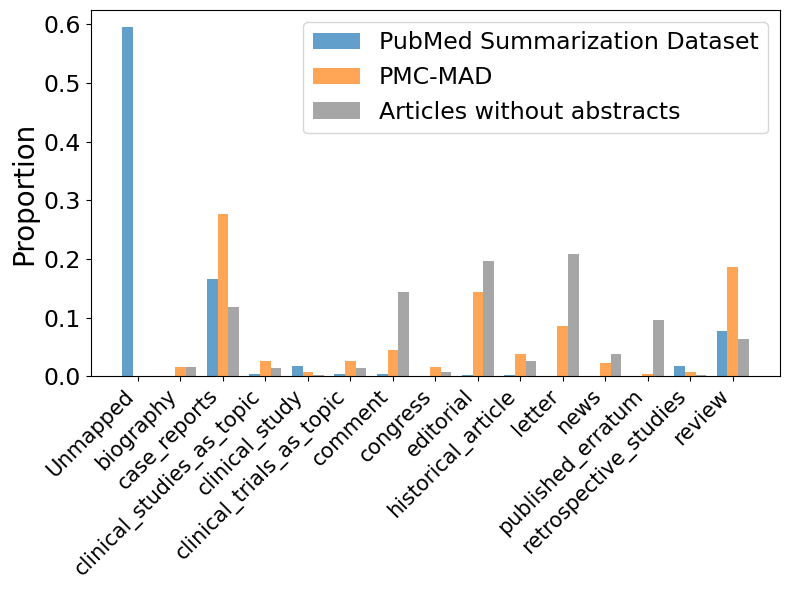}
    \caption{Publication type distribution of PMC-MAD, PubMed Summarization Dataset, 
and PubMed articles without abstracts.}
    \label{fig:PT_distribution}
\end{figure}

To evaluate the generalizability of our pipeline, we also tested DPR-BAG on the PubMed Summarization dataset \cite{pubmed_dataset}, hereafter PubMedSum. As shown in Figure \ref{fig:PT_distribution}, unlike PMC-MAD, which is stratified to reflect the publication type distribution of abstract-less PubMed manuscripts, the majority of articles in the PubMedSum could not be mapped to a specific publication type. This allows us to benchmark the performance of DPR-BAG against established standards in the biomedical domain and ensure that our findings are not limited to the specific characteristics of the PMC-MAD corpus.

\begin{figure}
    \centering
    \includegraphics[width=0.5\textwidth]{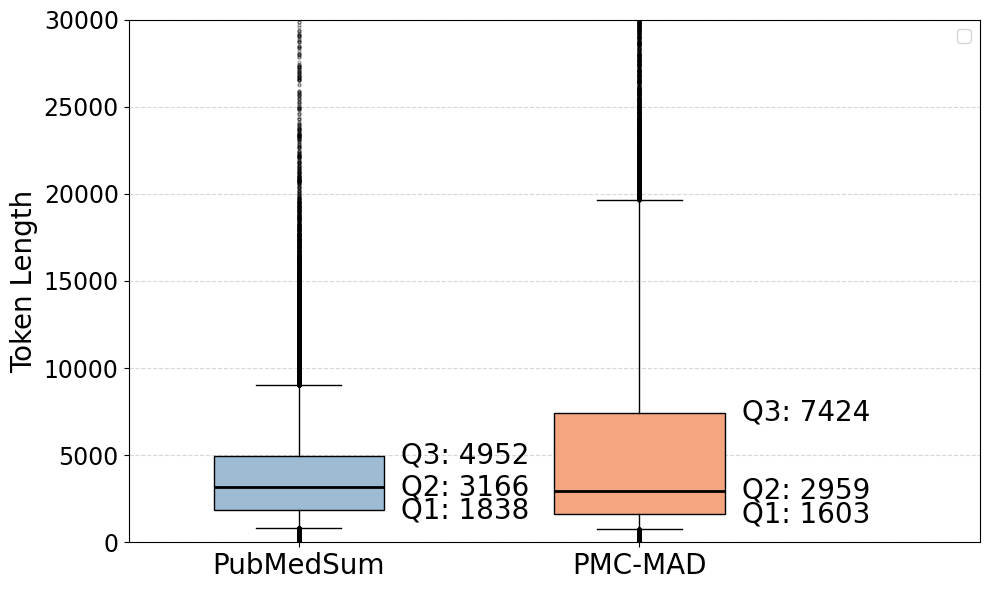} 
    \caption{Token Distribution Comparison}
    \label{fig:token_distribution}
\end{figure}

As illustrated in Figure \ref{fig:token_distribution}, the two datasets exhibit distinct token length profiles. PMC-MAD has a median document length of 2959 tokens (IQR: 1603--7424), while PubMedSum has a median of 3,166 tokens (IQR: 1838--4952). 

\begin{table*}[t]
\centering
\small
\setlength{\tabcolsep}{5pt}
\begin{tabular}{lccc|ccc}
\hline
 & \multicolumn{3}{c}{\textbf{Abstractiveness}} & \multicolumn{3}{c}{\textbf{Factuality}}\\
 \cline{2-4} \cline{5-7}
\textbf{Model} & \textbf{Bi-g} & \textbf{Tri-g} & \textbf{Dens.} & \textbf{AS} & \textbf{MC} & \textbf{SummaC}\\
\hline
Original abstract & 0.478 & 0.680 & 4.917 & -- & -- & --\\
\hline
LED-Arxiv (base) & 0.121 (-0.358) & 0.205 (-0.475) & 25.146 (+20.228) & 0.755 & \textbf{0.872} & \textbf{0.712}\\
LED-Pubmed (base) & 0.156 (-0.322) & 0.260 (-0.420) & 19.565 (+14.648) & 0.640 & 0.807 & 0.661\\
LongT5 (base) & 0.157 (-0.322) & 0.271 (-0.409) & 15.827 (+10.910) & 0.605 & 0.826 & 0.645\\
\hline
DPR-BAG (BC) & 0.507 (+0.029) & 0.733 (+0.053) & 2.966 (-1.951) & 0.765 & 0.794 & 0.414\\
DPR-BAG (SI+CoT) & \textbf{0.503 (+0.024)} & \textbf{0.723 (+0.043)} & \textbf{3.153 (-1.765)} & \textbf{0.772} & 0.787 & 0.411\\
\hline
\end{tabular}
\caption{Performance comparison on the PubMed Summarization dataset across primary evaluation dimensions. Conventions and abbreviations as in Table~\ref{tab:eval_primary}}
\label{tab:eval_pubmed_sum_primary}
\end{table*}

\begin{table*}[t]
\centering
\small
\setlength{\tabcolsep}{5pt}
\begin{tabular}{lccc|cccc}
\hline
 & \multicolumn{3}{c}{\textbf{Semantic Alignment}} & \multicolumn{4}{c}{\textbf{Supporting}}\\
 \cline{2-4} \cline{5-8}
\textbf{Model} & \textbf{BS} & \textbf{SENT} & \textbf{FOCUS$\downarrow$} & \textbf{R-L} & \textbf{U-R} & \textbf{Cov.} & \textbf{Comp.}\\
\hline
Original abstract & -- & -- & -- & -- & -- & 0.878 & 16.084\\
\hline
LED-Arxiv (base) & 0.641 & 0.913 & 1.398 & 0.236 & 0.348 & 0.968 (+0.091) & \textbf{17.739 (+1.655)}\\
LED-Pubmed (base) & 0.664 & \textbf{0.929} & \textbf{1.100} & \textbf{0.272} & \textbf{0.415} & 0.960 (+0.082) & 14.235 (-1.849)\\
LongT5 (base) & \textbf{0.664} & 0.894 & 1.585 & 0.265 & 0.345 & 0.962 (+0.085) & 25.643 (+9.560)\\
\hline
DPR-BAG (BC) & 0.651 & 0.814 & 2.116 & 0.192 & 0.217 & 0.876 (-0.002) & 43.962 (+27.878)\\
DPR-BAG (SI+CoT) & 0.652 & 0.839 & 1.927 & 0.197 & 0.245 & \textbf{0.878 (+0.000)} & 40.068 (+23.984)\\
\hline
\end{tabular}
\caption{Performance comparison on the PubMed Summarization dataset across semantic alignment and supporting metrics. Conventions and abbreviations as in Table~\ref{tab:eval_secondary}}
\label{tab:eval_pubmed_sum_secondary}
\end{table*}

Table \ref{tab:eval_pubmed_sum_primary} and Table \ref{tab:eval_pubmed_sum_secondary} present results on the PubMed Summarization dataset for the two best PMC-MAD configurations: BC (without entity guidance) and SI+CoT. Consistent with PMC-MAD findings, DPR-BAG underperforms all baseline models on ROUGE, UMLS Recall, DiscoScore, and SummaC, while achieving better coverage, density, and novel n-gram scores closer to the original human-written abstract. Both configurations outperform all baselines on AlignScore, confirming that the abstractiveness gains do not compromise factuality.

However, DPR-BAG's MiniCheck advantage on PMC-MAD does not transfer here, attributable to PubMedSum's inherent higher abstractiveness (lower density, higher novelty), which penalizes NLI-based metrics (MiniCheck) even when factual content is preserved — as evidenced by DPR-BAG's consistently superior AlignScore across both datasets.

Unlike on PMC-MAD where SI+CoT only improved BERTScore, on PubMedSum SI+CoT significantly improves AlignScore, ROUGE-L, UMLS Recall, DiscoScore, and Compression while showing no significant degradation on factuality (MiniCheck, SummaC) or abstractiveness. This suggests that CoT's benefit varies with document characteristics, potentially driven by publication-type distribution differences.

\begin{table*}[t]
\centering
\small
\setlength{\tabcolsep}{5pt}
\begin{tabular}{lccc|ccc}
\hline
 & \multicolumn{3}{c}{\textbf{Abstractiveness}} & \multicolumn{3}{c}{\textbf{Factuality}}\\
 \cline{2-4} \cline{5-7}
\textbf{Backbone} & \textbf{Bi-g} & \textbf{Tri-g} & \textbf{Dens.} & \textbf{AS} & \textbf{MC} & \textbf{SummaC}\\
\hline
Original abstract & 0.495 & 0.663 & 6.650 & -- & -- & --\\
\hline
Llama 3.2:3B & 0.397 (-0.098) & \textbf{0.605 (-0.058)} & \textbf{4.690 (-1.960)} & \textbf{0.762} & \textbf{0.890} & \textbf{0.570}\\
Qwen 2.5:7B & 0.594 (+0.099) & 0.806 (+0.143) & 2.380 (-4.270) & 0.635 & 0.793 & 0.432\\
Qwen 2.5:14B & \textbf{0.568 (+0.073)} & 0.778 (+0.115) & 2.651 (-3.999) & 0.647 & 0.812 & 0.443\\
\hline
\end{tabular}
\caption{Scale-up comparison on primary evaluation dimensions. All configurations use the BC prompt without entity guidance. Conventions and abbreviations as in Table~\ref{tab:eval_primary}}
\label{tab:scaling_primary}
\end{table*}

\begin{table*}[t]
\centering
\small
\setlength{\tabcolsep}{5pt}
\begin{tabular}{lccc|cccc}
\hline
 & \multicolumn{3}{c}{\textbf{Semantic Alignment}} & \multicolumn{4}{c}{\textbf{Supporting}}\\
 \cline{2-4} \cline{5-8}
\textbf{Backbone} & \textbf{BS} & \textbf{SENT} & \textbf{FOCUS$\downarrow$} & \textbf{R-L} & \textbf{U-R} & \textbf{Cov.} & \textbf{Comp.}\\
\hline
Original abstract & -- & -- & -- & -- & -- & 0.743 & 23.827\\
\hline
Llama 3.2:3B & 0.617 & 0.868 & 0.828 & 0.183 & 0.266 & 0.894 (+0.151) & 50.037 (+26.210)\\
Qwen 2.5:7B & 0.631 & \textbf{0.876} & 0.670 & 0.177 & 0.279 & \textbf{0.846 (+0.103)} & 32.273 (+8.446)\\
Qwen 2.5:14B & \textbf{0.633} & 0.866 & \textbf{0.596} & \textbf{0.189} & \textbf{0.337} & 0.857 (+0.114) & \textbf{23.234 (-0.593)}\\
\hline
\end{tabular}
\caption{Scale-up comparison on semantic alignment and supporting metrics. Conventions and abbreviations as in Table~\ref{tab:eval_secondary}}
\label{tab:scaling_secondary}
\end{table*}

\section{Effect of Backbone Model Size}
\label{appendix:scale-up}
To investigate whether scaling up the backbone LLM improves generation quality, we evaluated \texttt{Qwen2.5:7B} and \texttt{Qwen2.5:14B} under the BC prompt without entity guidance. As the Llama3.2 series does not offer a model beyond 3B, we use Qwen 2.5 for the scaling analysis. 

As shown in Table \ref{tab:scaling_primary} and \ref{tab:scaling_secondary}, Qwen2.5:14B improves BERTScore, UMLS Recall, and DS-Focus relative to the 3B backbone, recording the best DS-Focus, and UMLS Recall among all DPR configurations, alongside a compression rate most closely aligned with the original abstracts. However, both larger backbones exhibit excessively higher abstractiveness (lower Density, higher Novel n-grams) and declining factuality (SummaC, AlignScore and MiniCheck) relative to the 3B backbone, suggesting that larger models promote more abstractive phrasing and precise semantic alignment of key noun foci at the cost of reduced factual overlap with the source.

\section{BC Entity Guidance Ablations}
\label{appendix:bc_entity_guidance}

To isolate the effect of entity guidance from prompt complexity, 
we additionally apply TR-UMLS to the BC prompt and compare 
against BC without entity guidance.

We integrate TR-UMLS with top-5 UMLS entities into the BC prompt 
(BC+Top-5). As shown in Table~\ref{tab:bc_trumls}, BC+Top-5 
significantly degrades factuality relative to BC ($p<0.001$) and 
decreases UMLS Recall ($p<0.001$), indicating that the injected UMLS terms fail to anchor concept reproduction in the output. These results confirm that TR-UMLS entity guidance remains ineffective in this zero-shot setting regardless of the underlying prompt strategy. Significance details are reported in Appendix~\ref{appendix:sig_umls}.

\begin{table*}[h]
\centering
\small
\setlength{\tabcolsep}{5pt}
\begin{tabular}{lcc|cc|cc|c}
\hline
 & \multicolumn{2}{c}{\textbf{Abstractiveness}} & \multicolumn{2}{c}{\textbf{Factuality}} & \multicolumn{2}{c}{\textbf{Semantic Alignment}} & \\
 \cline{2-3} \cline{4-5} \cline{6-7} \cline{8-8}
\textbf{Config} & \textbf{Tri-g} & \textbf{Dens.} & \textbf{AS} & \textbf{MC} & \textbf{BS} & \textbf{FOCUS$\downarrow$} & \textbf{U-R}\\
\hline
BC & \textbf{0.605 (-0.058)} & \textbf{4.690 (-1.960)} & \textbf{0.762} & \textbf{0.890} & 0.617 & \textbf{0.828} & \textbf{0.266}\\
\hline
BC+Top-5 & 0.733 (+0.070) & 2.994 (-3.656) & 0.637 & 0.798 & 0.639 & 0.931 & 0.238\\
\hline
\end{tabular}
\caption{Ablation of TR-UMLS entity guidance applied to the BC prompt. 
(Abbreviations as in Table~\ref{tab:eval_primary} and Table~\ref{tab:eval_secondary})}
\label{tab:bc_trumls}
\end{table*}

\section{Evaluation Metric Details}
\label{appendix:metrics}
\subsection{Factual Consistency (AlignScore, MiniCheck and SummaC)}
\paragraph{AlignScore \cite{alignscore}:} AlignScore trains a unified alignment model on 4.7M examples spanning 7 tasks---NLI, fact verification, paraphrase, semantic textual similarity, question answering, information retrieval, and summarization---producing a single factual consistency score. At inference time, the source document is split into overlapping chunks of approximately 350 tokens, and each sentence in the generated abstract is evaluated against all chunks; the highest alignment score per sentence is averaged to yield the final score. We use the base variant (RoBERTa-base, 125M parameters).

\paragraph{MiniCheck \cite{minicheck}:} MiniCheck trains a small fact-checking model on synthetically constructed data generated by GPT-4, where each instance is designed to require verifying multiple atomic facts against multi-sentence evidence. It produces a binary supported/unsupported prediction per sentence. In our evaluation, each sentence in the generated abstract is treated as an individual claim and verified against the source full-text document. We use the Flan-T5-Large variant (770M parameters).

\paragraph{SummaC \cite{summaC}:} SummaC segments the source document into individual sentences and scores each generated sentence by aggregating sentence-level NLI entailment probabilities against all source sentences. The SummaC$_\text{Conv}$ variant learns a convolutional layer over the full distribution of these entailment scores, rather than relying only on the maximum, making it more robust to outliers. We use the \texttt{SummaCConv} implementation with a VitaminC-trained NLI backbone and sentence-level granularity.

\subsection{DiscoScore (DS\_SENT\_NN \& DS\_FOCUS\_NN)} 
DiscoScore\cite{discoscore} evaluates discourse coherence by modeling focus transitions across sentences. In this work, we use nouns (NN) as the focus, one of several focus choices supported by the metric, to compare the discourse coherence between the reference and generated abstracts.

\paragraph{DS\_SENT\_NN:} Constructs a sentence graph where edges are drawn between any two sentences (not just adjacent ones) that share at least one noun focus. Edge weights are inversely proportional to the distance between the two sentences ($1/(j-i)$), capturing both local and long-range coherence. Sentence embeddings are then aggregated according to this graph structure, and the cosine similarity between the resulting graph-level embeddings of the generated and reference texts is used as the final score.

\paragraph{DS\_FOCUS\_NN:} Measures how closely the frequency and semantics of shared noun foci match between the generated and reference texts. For each focus shared by both texts, it computes the distance between their embeddings (derived by summing the contextualized token embeddings of all associated tokens), and averages these distances as the final score.

\subsection{UMLS Recall}
This metric measures how well the generated abstract preserves biomedical concepts present in the reference abstract. We extract unique UMLS concepts from both the original abstract ($C_{\text{ref}}$) and the generated abstract ($C_{\text{gen}}$) using a biomedical entity linker (scispaCy, \citealp{scispacy}) with the UMLS knowledge base). UMLS Recall is then computed as:
\begin{equation}
    \text{UMLS\ Recall} = \frac{|C_{\text{gen}} \cap C_{\text{ref}}|}{|C_{\text{ref}}|}
\end{equation}

\paragraph{Newsroom \cite{newsroom}: Coverage, Density, and Compression: }
We adopt the extractive fragment measures from \citet{newsroom}. Given a summary $S$ and source document $D$, extractive fragments are the set of longest common substrings shared between $S$ and $D$. \textbf{Coverage} measures the proportion of summary tokens belonging to an extractive fragment. \textbf{Density} measures the average length of the extractive fragments, reflecting how verbatim the summary is. \textbf{Compression} is the ratio of source document length to summary length.

\paragraph{N-gram Novelty}
Following \citet{novelty}, we compute the proportion of bigrams and trigrams in the generated abstract that do not appear in the source full-text document. Higher novelty indicates greater abstractive synthesis relative to the source.

\section{Statistical Significance Analysis}
\label{appendix:significance_testing}
\subsection{Method}
For each comparison between two configurations $A$ and $B$, we align per-document scores by article ID and resample with replacement over 10,000 iterations. For metrics where higher (or lower) scores indicate better performance (AlignScore, MiniCheck, SummaC, BERTScore,   DiscoScore variants, ROUGE-L, UMLS Recall), we test $H_0\!: \mathbb{E}[s_A] = \mathbb{E}[s_B]$ on raw scores. For abstractiveness metrics where the goal is proximity to the 
reference distribution (Bigram novelty, Trigram novelty, Density, Coverage, Compression), we instead test $H_0\!: \mathbb{E}[\,|s_A - s_{\text{ref}}|\,] = \mathbb{E}[\,|s_B - s_{\text{ref}}|\,]$, where $s_{\text{ref}}$ is the corresponding score of the human-written abstract on the same document. Under this formulation, a negative difference indicates that $A$ is closer to the reference than $B$. All tests are two-sided, with significance markers $^{*}$ ($p<0.05$), $^{**}$ ($p<0.01$), and $^{***}$ ($p<0.001$). Comparisons not marked are not significant at $p<0.05$. The per-document scores used for testing are extracted from the same evaluation pipeline that produces the average numbers in Section \ref{sec:results}; only articles with valid scores from both configurations are included in each pairwise test.

In all tables below, mean differences are computed as $A\,-\,B$. Arrows next to metric names indicate whether higher ($\uparrow$) or lower ($\downarrow$) values are preferred.

\subsection{DPR-BAG vs.\ Baselines (PMC-MAD)}
\label{appendix:sig_baselines}
Table~\ref{tab:sig_baselines_pretrained}, Table~\ref{tab:sig_baselines_other}, and Table~\ref{tab:sig_baselines_longt5ft} report significance for the comparisons between DPR-BAG (BC) and the baseline models on PMC-MAD.
 
\begin{table}[h]
\centering
\small
\setlength{\tabcolsep}{4pt}
\begin{tabular}{lcc}
\toprule
Metric & vs.\ LED-arXiv & vs.\ LED-PubMed  \\
\midrule
Bi-g$\downarrow$  & $-0.152^{***}$ & $-0.131^{***}$ \\
Tri-g$\downarrow$ & $-0.217^{***}$ & $-0.187^{***}$ \\
Dens.$\downarrow$ & $-12.121^{***}$ & $-9.048^{***}$ \\
AS$\uparrow$      & $+0.041^{***}$ & $+0.095^{***}$ \\
MC$\uparrow$      & $+0.033^{***}$ & $+0.044^{***}$ \\
SummaC$\uparrow$  & $-0.167^{***}$ & $-0.124^{***}$ \\
\midrule
BS$\uparrow$      & $-0.019^{***}$ & $-0.036^{***}$ \\
SENT$\uparrow$    & $-0.039^{***}$ & $-0.040^{***}$ \\
FOCUS$\downarrow$ & $+0.126^{***}$ & $+0.158^{***}$ \\
R-L$\uparrow$     & $-0.052^{***}$ & $-0.080^{***}$ \\
U-R$\uparrow$     & $-0.075^{***}$ & $-0.089^{***}$ \\
Cov.$\downarrow$  & $-0.017^{***}$ & $-0.011^{***}$ \\
Comp.$\downarrow$ & $+20.499^{***}$ & $+20.720^{***}$ \\
\bottomrule
\end{tabular}
\caption{Mean differences (DPR-BAG minus baseline) between 
DPR-BAG (BC) and the pretrained LED baselines on PMC-MAD.}
\label{tab:sig_baselines_pretrained}
\end{table}
 
\begin{table}[h]
\centering
\small
\setlength{\tabcolsep}{4pt}
\begin{tabular}{lcc}
\toprule
Metric & vs.\ LED-PubMed-FT & vs.\ LongT5 \\
\midrule
Bi-g$\downarrow$  & $-0.032^{***}$ & $-0.128^{***}$ \\
Tri-g$\downarrow$ & $-0.042^{***}$ & $-0.170^{***}$ \\
Dens.$\downarrow$ & $-2.383^{***}$  & $-5.565^{***}$ \\
AS$\uparrow$      & $+0.251^{***}$ & $+0.093^{***}$ \\
MC$\uparrow$      & $+0.236^{***}$ & $+0.013^{***}$ \\
SummaC$\uparrow$  & $+0.067^{***}$ & $-0.142^{***}$ \\
\midrule
BS$\uparrow$      & $-0.018^{***}$ & $-0.048^{***}$ \\
SENT$\uparrow$    & $-0.000$       & $-0.008^{***}$ \\
FOCUS$\downarrow$ & $+0.033$       & $-0.036$ \\
R-L$\uparrow$     & $-0.082^{***}$ & $-0.093^{***}$ \\
U-R$\uparrow$     & $-0.099^{***}$ & $-0.047^{***}$ \\
Cov.$\downarrow$  & $+0.052^{***}$ & $-0.015^{***}$ \\
Comp.$\downarrow$ & $+22.307^{***}$ & $+6.213^{*}$ \\
\bottomrule
\end{tabular}
\caption{Mean differences (DPR-BAG minus baseline) between 
DPR-BAG (BC) and the fine-tuned LED-PubMed and LongT5 baselines.}
\label{tab:sig_baselines_other}
\end{table}

\begin{table}[h]
\centering
\small
\setlength{\tabcolsep}{4pt}
\begin{tabular}{lc}
\toprule
Metric & vs.\ LongT5-FT \\
\midrule
Bi-g$\downarrow$  & $-0.059^{***}$ \\
Tri-g$\downarrow$ & $-0.081^{***}$ \\
Dens.$\downarrow$ & $-3.048^{***}$ \\
AS$\uparrow$      & $+0.221^{***}$ \\
MC$\uparrow$      & $+0.164^{***}$ \\
SummaC$\uparrow$  & $+0.005$ \\
\midrule
BS$\uparrow$      & $-0.034^{***}$ \\
SENT$\uparrow$    & $-0.027^{***}$ \\
FOCUS$\downarrow$ & $+0.111^{***}$ \\
R-L$\uparrow$     & $-0.095^{***}$ \\
U-R$\uparrow$     & $-0.097^{***}$ \\
Cov.$\downarrow$  & $+0.021^{***}$ \\
Comp.$\downarrow$ & $+19.467^{***}$ \\
\bottomrule
\end{tabular}
\caption{Mean differences (DPR-BAG minus baseline) between 
DPR-BAG (BC) and the fine-tuned LongT5 baseline.}
\label{tab:sig_baselines_longt5ft}
\end{table}

\subsection{Splitting Strategies}
\label{appendix:sig_splitting}
 
Table~\ref{tab:sig_splitting} reports significance for the comparisons between FS and other splitting strategies.
 
\begin{table}[h]
\centering
\small
\setlength{\tabcolsep}{4pt}
\begin{tabular}{lcc}
\toprule
Metric & vs.\ NS & vs.\ SH \\
\midrule
Bi-g$\downarrow$  & $+0.005^{*}$   & $+0.045^{***}$  \\
Tri-g$\downarrow$ & $+0.006^{*}$   & $+0.046^{***}$  \\
Dens.$\downarrow$ & $+0.150^{**}$  & $+0.698^{***}$  \\
AS$\uparrow$      & $+0.006^{*}$   & $+0.126^{***}$  \\
MC$\uparrow$      & $-0.003$       & $+0.099^{***}$  \\
SummaC$\uparrow$  & $+0.003$       & $+0.141^{***}$  \\
\midrule
BS$\uparrow$      & $+0.001$       & $-0.019^{***}$  \\
SENT$\uparrow$    & $+0.004^{*}$   & $+0.015^{***}$  \\
FOCUS$\downarrow$ & $-0.033$       & $-0.098^{***}$  \\
R-L$\uparrow$     & $+0.002^{*}$   & $+0.004^{***}$  \\
U-R$\uparrow$     & $+0.004$       & $+0.037^{***}$  \\
Cov.$\downarrow$  & $+0.000$       & $+0.057^{***}$  \\
Comp.$\downarrow$ & $+1.875$        & $+1.897$       \\
\bottomrule
\end{tabular}
\caption{Mean differences between FS and other splitting strategies.}
\label{tab:sig_splitting}
\end{table}

\subsection{Prompting Strategies}
\label{appendix:sig_prompting}
 
Table~\ref{tab:sig_prompting} reports significance for the 
prompting strategy ablation.
 
\begin{table}[h]
\centering
\small
\setlength{\tabcolsep}{4pt}
\begin{tabular}{lccc}
\toprule
Metric & BC vs.\ DI & BC vs.\ SI & DI vs.\ SI \\
\midrule
Bi-g$\downarrow$  & $+0.041^{***}$ & $+0.037^{***}$ & $-0.004^{**}$ \\
Tri-g$\downarrow$ & $+0.045^{***}$ & $+0.040^{***}$ & $-0.005^{***}$ \\
Dens.$\downarrow$ & $+0.662^{***}$ & $+0.616^{***}$  & $-0.045^{*}$ \\
AS$\uparrow$      & $+0.121^{***}$ & $+0.132^{***}$ & $+0.012^{***}$ \\
MC$\uparrow$      & $+0.084^{***}$ & $+0.086^{***}$ & $+0.002$ \\
SummaC$\uparrow$  & $+0.133^{***}$ & $+0.139^{***}$ & $+0.006^{**}$ \\
\midrule
BS$\uparrow$      & $-0.022^{***}$ & $-0.019^{***}$ & $+0.003^{***}$ \\
SENT$\uparrow$    & $+0.007^{***}$ & $+0.004^{**}$  & $-0.003$ \\
FOCUS$\downarrow$ & $-0.065^{**}$  & $-0.013$       & $+0.050^{**}$ \\
R-L$\uparrow$     & $+0.001$       & $+0.001$       & $+0.001$ \\
U-R$\uparrow$     & $+0.018^{***}$ & $+0.014^{***}$ & $-0.004^{*}$ \\
Cov.$\downarrow$  & $+0.052^{***}$ & $+0.061^{***}$ & $+0.009^{***}$ \\
Comp.$\downarrow$ & $+3.895$        & $+5.627$        & $+1.732^{*}$ \\
\bottomrule
\end{tabular}
\caption{Mean differences across prompting strategies.}
\label{tab:sig_prompting}
\end{table}

\subsection{TR-UMLS Entity Guidance}
\label{appendix:sig_umls}
 
Table~\ref{tab:sig_umls} and Table~\ref{tab:sig_umls_vs_bc} report significance for the TR-UMLS entity guidance ablation.
 
\begin{table}[h]
\centering
\small
\setlength{\tabcolsep}{4pt}
\begin{tabular}{lcc}
\toprule
Metric & vs.\ DI+Top-5 & vs.\ DI+Top-10 \\
\midrule
Bi-g$\downarrow$  & $+0.000$       & $-0.021^{***}$ \\
Tri-g$\downarrow$ & $-0.000$       & $-0.030^{***}$ \\
Dens.$\downarrow$ & $-0.011$       & $-0.137^{***}$ \\
AS$\uparrow$      & $-0.003$       & $-0.002$ \\
MC$\uparrow$      & $-0.001$       & $-0.006$ \\
SummaC$\uparrow$  & $+0.000$       & $-0.002$ \\
\midrule
BS$\uparrow$      & $+0.001$       & $+0.043^{***}$ \\
SENT$\uparrow$    & $+0.002$       & $+0.001$ \\
FOCUS$\downarrow$ & $-0.048^{**}$  & $-0.026$ \\
R-L$\uparrow$     & $-0.001$       & $+0.011^{***}$ \\
U-R$\uparrow$     & $+0.002$       & $+0.016^{***}$ \\
Cov.$\downarrow$  & $-0.002^{**}$  & $-0.043^{***}$ \\
Comp.$\downarrow$ & $+0.395$        & $-0.069$ \\
\bottomrule
\end{tabular}
\caption{Mean differences between DI and TR-UMLS-augmented DI.}
\label{tab:sig_umls}
\end{table}

\begin{table}[h]
\centering
\small
\setlength{\tabcolsep}{4pt}
\begin{tabular}{lccc}
\toprule
Metric & vs.\ BC+Top-5 & vs.\ DI+Top-5 & vs.\ DI+Top-10 \\
\midrule
Bi-g$\downarrow$  & $+0.040^{***}$ & $+0.041^{***}$ & $+0.021^{***}$ \\
Tri-g$\downarrow$ & $+0.044^{***}$ & $+0.045^{***}$ & $+0.016^{***}$ \\
Dens.$\downarrow$ & $+0.685^{***}$& $+0.650^{***}$  & $+0.525^{***}$ \\
AS$\uparrow$      & $+0.126^{***}$& $+0.118^{***}$ & $+0.119^{***}$ \\
MC$\uparrow$      & $+0.092^{***}$& $+0.083^{***}$ & $+0.078^{***}$ \\
SummaC$\uparrow$  & $+0.134^{***}$& $+0.133^{***}$ & $+0.130^{***}$ \\
\midrule
BS$\uparrow$      & $-0.022^{***}$ & $-0.021^{***}$ & $+0.022^{***}$ \\
SENT$\uparrow$    & $+0.014^{***}$ & $+0.008^{***}$ & $+0.008^{***}$ \\
FOCUS$\downarrow$ & $-0.107^{***}$ & $-0.113^{***}$ & $-0.087^{***}$ \\
R-L$\uparrow$     & $+0.002$ & $+0.000$       & $+0.012^{***}$ \\
U-R$\uparrow$     & $+0.028^{***}$ & $+0.021^{***}$ & $+0.035^{***}$ \\
Cov.$\downarrow$  & $+0.051^{***}$ & $+0.050^{***}$ & $+0.009^{***}$ \\
Comp.$\downarrow$ & $+3.045$ & $+4.291$        & $+3.827$ \\
\bottomrule
\end{tabular}
\caption{Mean differences between BC and TR-UMLS-augmented variants.}
\label{tab:sig_umls_vs_bc}
\end{table}

\subsection{Chain-of-Thought Guidance}
\label{appendix:sig_cot}
 
Table~\ref{tab:sig_cot} reports significance for the SI+CoT 
ablation.
 
\begin{table}[h]
\centering
\small
\setlength{\tabcolsep}{4pt}
\begin{tabular}{lcc}
\toprule
Metric & BC vs.\ SI+CoT & SI vs.\ SI+CoT \\
\midrule
Bi-g$\downarrow$  & $+0.032^{***}$ & $-0.005^{**}$ \\
Tri-g$\downarrow$ & $+0.037^{***}$ & $-0.003$ \\
Dens.$\downarrow$ & $+0.591^{***}$  & $-0.030$ \\
AS$\uparrow$      & $+0.166^{***}$ & $+0.034^{***}$ \\
MC$\uparrow$      & $+0.098^{***}$ & $+0.012^{**}$ \\
SummaC$\uparrow$  & $+0.158^{***}$ & $+0.019^{***}$ \\
\midrule
BS$\uparrow$      & $-0.015^{***}$ & $+0.005^{***}$ \\
SENT$\uparrow$    & $-0.000$       & $-0.004^{***}$ \\
FOCUS$\downarrow$ & $+0.031$       & $+0.042^{*}$ \\
R-L$\uparrow$     & $+0.005^{***}$ & $+0.004^{***}$ \\
U-R$\uparrow$     & $+0.017^{***}$ & $+0.003$ \\
Cov.$\downarrow$  & $+0.070^{***}$ & $+0.009^{***}$ \\
Comp.$\downarrow$ & $+7.488^{*}$    & $+1.860^{***}$ \\
\bottomrule
\end{tabular}
\caption{Mean differences for the SI+CoT ablation.}
\label{tab:sig_cot}
\end{table}
 
\subsection{PubMedSum Generalization}
\label{appendix:sig_pubmedsum}
 
Table~\ref{tab:sig_pubmedsum_bs} reports significance for 
DPR-BAG (BC) against the baseline models on PubMedSum, and 
Table~\ref{tab:sig_pubmedsum_cot} compares BC against SI+CoT 
on the same dataset.
 
\begin{table}[h]
\centering
\small
\setlength{\tabcolsep}{4pt}
\begin{tabular}{lccc}
\toprule
Metric & vs.\ LED-arXiv & vs.\ LED-PubMed & vs.\ LongT5 \\
\midrule
Bi-g$\downarrow$  & $-0.226^{***}$ & $-0.190^{***}$ & $-0.190^{***}$ \\
Tri-g$\downarrow$ & $-0.333^{***}$ & $-0.279^{***}$ & $-0.269^{***}$ \\
Dens.$\downarrow$ & $-17.806^{***}$ & $-12.261^{***}$ & $-8.710^{***}$ \\
AS$\uparrow$      & $+0.010^{**}$  & $+0.125^{***}$ & $+0.160^{***}$ \\
MC$\uparrow$      & $-0.078^{***}$ & $-0.013^{**}$  & $-0.032^{***}$ \\
SummaC$\uparrow$  & $-0.299^{***}$ & $-0.248^{***}$ & $-0.231^{***}$ \\
\midrule
BS$\uparrow$      & $+0.011^{***}$ & $-0.012^{***}$ & $-0.013^{***}$ \\
SENT$\uparrow$    & $-0.099^{***}$ & $-0.115^{***}$ & $-0.080^{***}$ \\
FOCUS$\downarrow$ & $+0.719^{***}$ & $+1.015^{***}$ & $+0.532^{***}$ \\
R-L$\uparrow$     & $-0.044^{***}$ & $-0.081^{***}$ & $-0.073^{***}$ \\
U-R$\uparrow$     & $-0.131^{***}$ & $-0.198^{***}$ & $-0.127^{***}$ \\
Cov.$\downarrow$  & $-0.035^{***}$ & $-0.024^{***}$ & $-0.027^{***}$ \\
Comp.$\downarrow$ & $+21.934^{***}$ & $+23.572^{***}$ & $+16.835^{***}$ \\
\bottomrule
\end{tabular}
\caption{Mean differences (DPR-BAG minus baseline) between 
DPR-BAG (BC) and baseline models on PubMedSum.}
\label{tab:sig_pubmedsum_bs}
\end{table}
 
\begin{table}[h]
\centering
\small
\setlength{\tabcolsep}{4pt}
\begin{tabular}{lc}
\toprule
Metric & BC vs.\ SI+CoT \\
\midrule
Bi-g$\downarrow$  & $-0.002$ \\
Tri-g$\downarrow$ & $-0.001$ \\
Dens.$\downarrow$ & $-0.016$ \\
AS$\uparrow$      & $-0.007^{*}$ \\
MC$\uparrow$      & $+0.007$ \\
SummaC$\uparrow$  & $+0.003$ \\
\midrule
BS$\uparrow$      & $-0.000$ \\
SENT$\uparrow$    & $-0.025^{***}$ \\
FOCUS$\downarrow$ & $+0.189^{***}$ \\
R-L$\uparrow$     & $-0.006^{***}$ \\
U-R$\uparrow$     & $-0.028^{***}$ \\
Cov.$\downarrow$  & $+0.001$ \\
Comp.$\downarrow$ & $+2.999^{***}$ \\
\bottomrule
\end{tabular}
\caption{Mean differences between DPR-BAG (BC) and DPR-BAG 
(SI+CoT) on PubMedSum.}
\label{tab:sig_pubmedsum_cot}
\end{table}

\subsection{Backbone Scale}
\label{appendix:sig_scale}
 
Table~\ref{tab:sig_scale} reports significance for the backbone 
scaling experiments.
 
\begin{table}[h]
\centering
\small
\setlength{\tabcolsep}{4pt}
\begin{tabular}{lcc}
\toprule
Metric & vs.\ Qwen-7B & vs.\ Qwen-14B \\
\midrule
Bi-g$\downarrow$  & $+0.006$       & $+0.043^{***}$ \\
Tri-g$\downarrow$ & $-0.005$       & $+0.046^{***}$ \\
Dens.$\downarrow$ & $+0.519^{***}$  & $+0.736^{***}$  \\
AS$\uparrow$      & $+0.131^{***}$ & $+0.115^{***}$ \\
MC$\uparrow$      & $+0.098^{***}$ & $+0.078^{***}$ \\
SummaC$\uparrow$  & $+0.139^{***}$ & $+0.128^{***}$ \\
\midrule
BS$\uparrow$      & $+0.048^{***}$ & $-0.015^{***}$  \\
SENT$\uparrow$    & $-0.008^{***}$ & $+0.003$        \\
FOCUS$\downarrow$ & $+0.145^{***}$ & $+0.242^{***}$  \\
R-L$\uparrow$     & $+0.023^{***}$ & $-0.006^{***}$  \\
U-R$\uparrow$     & $+0.013^{***}$ & $-0.070^{***}$  \\
Cov.$\downarrow$  & $+0.012^{***}$ & $+0.068^{***}$  \\
Comp.$\downarrow$ & $+15.960^{***}$ & $+21.741^{***}$  \\
\bottomrule
\end{tabular}
\caption{Mean differences (Llama-3.2:3B minus Qwen) between Llama-3.2:3B and the scaled Qwen2.5 backbones (denoted Qwen-7B and Qwen-14B).}
\label{tab:sig_scale}
\end{table}

\section{Case Study: Qualitative Comparison of Generated Abstracts}
\label{appendix:case_study}

As shown in Table~\ref{tab:case-study}, both LED and LongT5 models exhibit severe verbatim copying from the source full text, with LongT5 further producing a lexical hallucination. In contrast, DPR-BAG demonstrates stronger topic identification and more abstractive generation for the BAG task, though it still occasionally exhibits entity relation confusion (e.g., misattributing target genes as lncRNAs).
\begin{table*}[t]
\small
\begin{tabularx}{\textwidth}{p{2.2cm}X}
\toprule
\textbf{Model} & \textbf{Generated Abstract (truncated)} \\
\midrule
Original Abstract & 
Long non-coding RNAs (lncRNAs) comprise a sizeable class of non-coding RNAs with a length of over 200 base pairs. Little is known about their biological function, although over 20,000 lncRNAs have been annotated in the human genome. Through a diverse range of mechanisms, their primary function is in the regulation of the transcription of protein-coding genes. lncrna transcriptional activation can result from a group of nucleus-retained and chromatin-associated lncrnas, which function as scaffolds in the cis/trans recruitment of transcription factors, co-activators or chromatin remodelers, and/or promoter enhancers. their synthesis is initiated by various processes including ceramide synthesis, release of intracellular ca2+ or acid-base balance disorders. prior to vesicle creation, selective cargo loading occurs in the endosomal sorting complex required for transport. participation of endosomal sorting proteins such as tetraspanins or specific sumoylated proteins required for transport has been indicated in research. the endosomal-sorting complex consists of four components, these induce the formation of multivesicular bodies and the induction of membrane deformation to form exosomes. nanovesicles could be formed inside multivesicular bodies to allow transport outside the cell or digestion in lysosomes.  the molecular content of exosomes is more heterogenic than its synthesis process, with different cargoes being examined inside vesicles with regard to the type or stage of cancers. this paper will review the importance of lncrnas as crucial molecular content of exosomes, indicating its involvement in tumour suppression, pro-tumorigenic events and the development of novel therapeutic approaches in the near future. further studies of their mechanisms of function are essential, as well as overcoming several challenges to gain a clearer insight to the approaches for the best clinical application. 
\\
\midrule
LED-Pubmed (FT) & 
The exosome is a type of \textcolor{orange}{extracellular vesicle involved in cell-to-cell communication and disease transmission. Exosomes can be found in urine, sperm, amniotic fluid and other biological fluids}, and have been implicated in various biological processes, such as cancer development, immune responses, invasion, invasion and invasion, cell proliferation, \textcolor{red}{invasion or invasion, migration and invasion of tumour cells}, as well as in the regulation of gene expression in cancer cells. the lncrna (non-coding rna) is a class of transcripts that do \textcolor{orange}{not encode proteins and whose length exceeds 200 bp} and \textcolor{orange}{therefore play a vital role in regulating signaling pathways and gene expression}, and therefore plays a pivotal role in cancer development and cancer-related diseases such as metastasis, angiogenesis and tumour cell invasion, and is involved in the formation of exosomal lipid bilayers, migration, proliferation and invasion in tumour microenvironment, which in turn are involved in cancer progression and progression, as evidenced by their involvement in the secretion of cytokines and growth factors, lipids or proteins from endosomes, and in the activation of the transcription factor hypoxia inducible factor (hif-$\beta$) pathway, among others, and their role in the induction of pre-metastatic niche formation in cancerous diseases, including lung cancer. the purpose of this review is to \textcolor{orange}{summarize the role of various exosomally}-derived \textcolor{red}{lnas} in carcinogenesis and in other biological processes and to highlight the importance of these molecules in cancer biology and cancer development. 
 \\
\midrule
LongT5 (Base) & 
Circulating long non-coding rnas are a class of transcripts that do not encode proteins and whose length exceeds 200 bp. \textcolor{orange}{They play a significant role in diverse cellular processes from enabling normal development to disease progression, as well as cell communication. A valuable example of a biological carrier that can transport abundant quantities of} \textcolor{red}{long non - codons} \textcolor{orange}{is that of the exosome. its role has been documented in both physiological conditions and pathological changes such as cancer development, where they are responsible for the regulation of various processes including immunosuppression, proliferation and induction of pre - metastatic niche formation. in this review, the role of} exosomal long non
  \\
\midrule
\textbf{DPR-BAG (BC)} & 
LncRNA has greater tissue specificity than mRNA, making it a potential diagnostic biomarker for specific tumour types. Circulating RNA in body fluids can resist ribonucleases, providing stability and potential as cancer prognostic markers. A study found two groups of long non-coding RNAs (lncRNAs) with opposite effects on cancer cells. Gas5 was shown to induce apoptosis and inhibit tumour growth, while Particl enhanced DNA methylation and gene silencing. Another lncRNA, CCND1, bound to a modulator and blocked transcription factors. \textcolor{red}{Exo1-4} lncRNAs increased recipient cell viability by interacting with proteins involved in metabolism and nucleosomal architecture. Exosomal long \textcolor{red}{non-coding RNAs(lncRNAs)} have shown potential as effective nanocarriers for regulating lncRNA expression in cancer cells, offering a promising tumour therapy approach.
\\
\bottomrule
\end{tabularx}
\caption{Qualitative comparison of generated abstracts for a sample article (PMCID: PMC6625196) from the PMC-MAD test set. \textcolor{orange}{Orange}: verbatim copy from source full text. \textcolor{red}{Red}: factual error.}
\label{tab:case-study}
\end{table*}

\end{document}